\documentclass{article}

\usepackage[final]{neurips_2023}

\usepackage[utf8]{inputenc} 
\usepackage[T1]{fontenc}    
\usepackage{hyperref}       
\usepackage{url}            
\usepackage{booktabs}       
\usepackage{amsfonts}       
\usepackage{nicefrac}       
\usepackage{microtype}      
\usepackage{xcolor}         

\usepackage[textsize=tiny]{todonotes}
\usepackage{inconsolata}
\usepackage{multirow}
\usepackage{multicol}
\usepackage{tabularx}
\usepackage{amsmath}
\usepackage{amssymb}
\usepackage{mathtools}
\usepackage{enumitem}
\usepackage{amsthm}
\usepackage{bbm}
\usepackage{comment}

\usepackage{caption}
\usepackage{subcaption}
\usepackage{wrapfig}

\theoremstyle{definition}

\newcommand{\STAB}[1]{\begin{tabular}{@{}c@{}}#1\end{tabular}}

\newcommand\ttsmall[1]{\texttt{\small #1}}
\newcommand{\satprob}{\mathcal{P}}
\newcommand{\theory}{\mathcal{T}}
\newcommand{\gsm}{\textsc{GSM}}
\newcommand{\spec}{\Phi}
\newcommand{\query}{Q}
\newcommand{\gsmsys}{\textsc{GSM-Sys}}
\newcommand{\algebra}{\textsc{Algebra}}
\newcommand{\lsat}{\textsc{LSAT}}
\newcommand{\clutrr}{\textsc{Clutrr}}
\newcommand{\colorobj}{\textsc{Color}}
\newcommand{\proofwriter}{\textsc{ProofWriter}}
\newcommand{\stregex}{\textsc{StRegex}}
\newcommand{\boardgame}{\textsc{BoardgameQA}}

\newcommand{\cotlm}{\textsc{CoT}}
\newcommand{\satlm}{\textsc{SatLM}}
\newcommand{\pallm}{\textsc{ProgLM}}

\title{\textsc{SatLM}: Satisfiability-Aided Language Models \\ Using Declarative Prompting}

\author{%
\bf Xi Ye\quad\quad Qiaochu Chen\quad\quad Isil Dillig\quad\quad Greg Durrett\\
  Department of Computer Science \\
  The University of Texas at Austin \\
  \texttt{\{xiye,qchen,isil,gdurrett\}@cs.utexas.edu} \\
  }

\begin{document}

\maketitle

\begin{abstract}


Prior  work has combined  chain-of-thought prompting in large language models (LLMs) with programmatic representations to perform effective and transparent reasoning.  While such an approach works well for tasks that only require forward reasoning (e.g., straightforward arithmetic), it is less effective for constraint solving problems that require more sophisticated planning and search.  In this paper, we propose a new \emph{satisfiability-aided language modeling} (\satlm{}) approach for improving the reasoning capabilities of LLMs. We use an LLM to generate a \emph{declarative task specification} rather than an imperative program and leverage an off-the-shelf automated theorem prover to derive the final answer. This approach has two key advantages. The declarative specification is closer to the problem description than the reasoning steps are, so the LLM can parse it out of the description more accurately. Furthermore, by offloading the actual reasoning task to an automated theorem prover, our approach can guarantee the correctness of the answer with respect to the parsed specification and avoid planning errors in the solving process.
We evaluate \satlm{} on 8 different datasets and show that it consistently outperforms program-aided LMs in the imperative paradigm.
In particular, \satlm{} outperforms program-aided LMs by 23\% on a challenging subset of the \gsm{} arithmetic reasoning dataset; \satlm{} also achieves a new SoTA on \lsat{} and \boardgame{}, surpassing previous models that are trained on the respective training sets.\footnote{Code available at \url{https://github.com/xiye17/SAT-LM}.}

\end{abstract}

\section{Introduction}

Using large language models (LLMs) to perform complex reasoning has been a central thrust of recent research~\citep{gpt3,palm,Rae2021ScalingLM,metaopt}. Techniques like scratchpads~\citep{scratch} or chain-of-thought prompting (CoT)~\citep{chain} enable LLMs to follow a sequence of reasoning steps before making a prediction.
This paradigm is effective on various multi-step reasoning tasks, especially those with fixed forward reasoning procedures~\citep{chain}, e.g., concatenating the last letters of several words. However, CoT prompting can fall short when scaling to problems that involve intensive computation~\citep{pal} or long sequences of reasoning steps~\citep{creswell2023selectioninference,saparov2023language,ribeiro2023street}.

\begin{figure}[t]
  \begin{center}
    \includegraphics[width=1.0\linewidth,trim=0 135 180 0,clip]{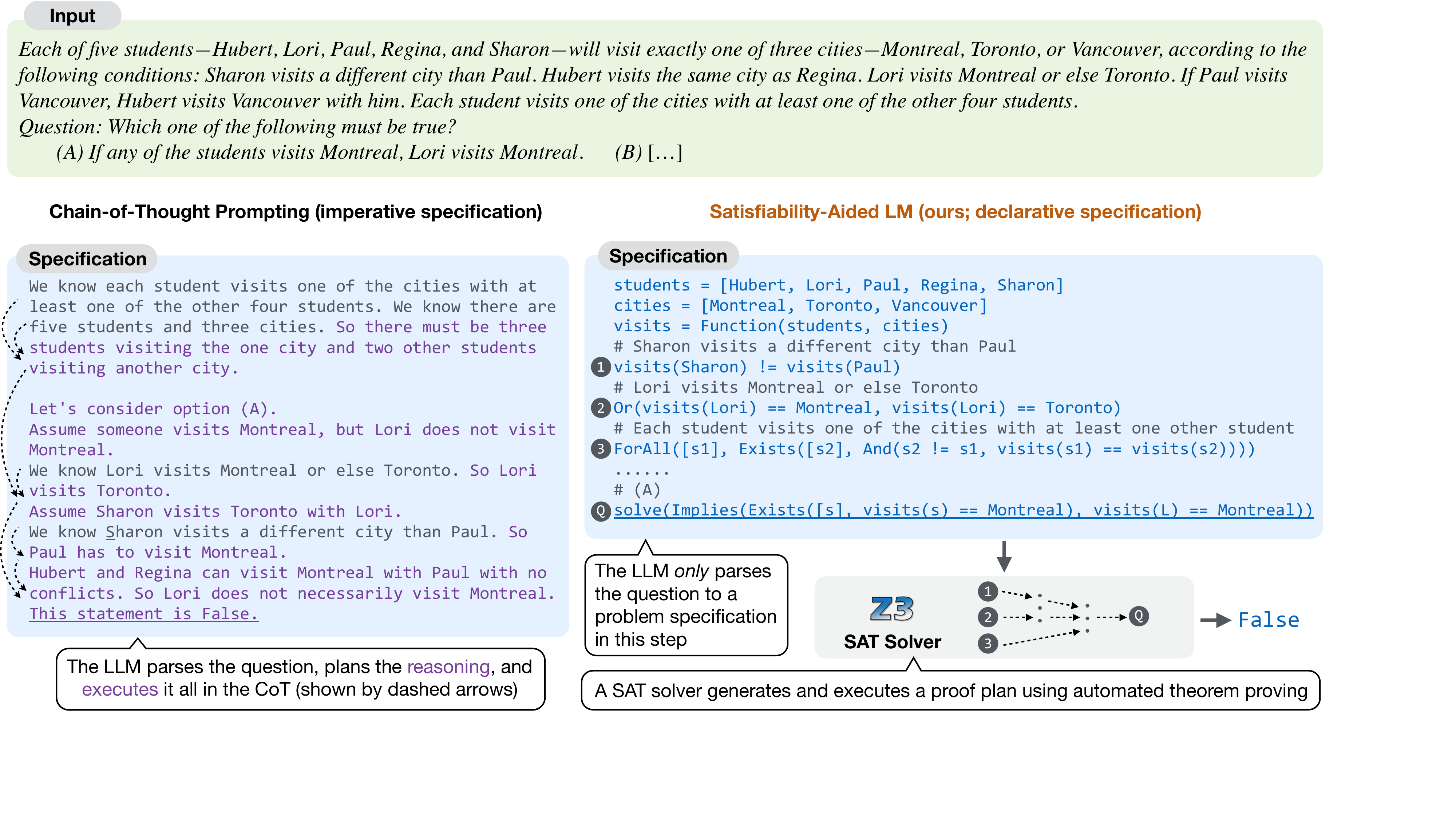}
  \end{center}
  \caption{Illustration of our Satisfiability-aided Language Modeling approach (right). We first parse an NL input into a declarative task specification (a set of logical constraints) using prompting (Section~\ref{sec:sat_prompt}), then use a SAT solver to solve the problem (Section~\ref{sec:sat_solver}). The chain-of-thought strategy in prior work (left) yields imperative reasoning processes.}

\label{fig:framework}
\end{figure}

%
Solving a complex reasoning problem involves three conceptual components: parsing a natural language description into a representation of the problem, deriving a plan to solve the problem, and executing that plan to obtain an answer.
Recent work on improving CoT prompting focuses on fixing \emph{execution errors} by augmenting LLMs with symbolic executors such as a Python interpreter, which leads to improved performance on arithmetic and symbolic reasoning tasks~\citep{pal,progcot,faithfulcot}. However, CoT prompting~\citep{chain,scratch} and its executor-augmented successors~\citep{pal,progcot,faithfulcot} are oriented towards \emph{imperative} solving procedures: a CoT or a program specifies the reasoning procedure as chained steps~\citep{chain,pal} in the order of execution. While this is effective for problems whose natural language already provides a suitably clear ``plan'' for the reasoning, it only leads to limited success for reasoning problems like in Figure~\ref{fig:framework} that do not outline such a plan~\citep{ribeiro2023street}. These problems often state a set of premises and constraints and ask questions that require sophisticated planning to deductively reason over the inputs, which is still challenging even for modern LLMs~\citep{Valmeekam2022LargeLM}.

Our work tackles both execution errors and, more importantly, \emph{planning errors}.
We propose SATisfiablity-aided Language Modeling (\satlm{}) using declarative prompting. The core idea is to cast a natural language (NL) reasoning problem as a satisfiability (SAT for short) problem. As shown in Figure~\ref{fig:framework} (right), given a problem in NL, we prompt an LLM to parse it into a SAT problem specification which consists of a set of logical formulas, then obtain the solution by invoking a SAT solver.\footnote{Here, we use SAT solver to refer to any automated reasoning tool for checking the satisfiability of formulas in formal logic. Hence, ``SAT solver'' in this paper also includes first-order theorem provers and SMT solvers.} 
The LLM is specialized towards understanding the preconditions stated in the problem, while the solver is leveraged to plan out the reasoning procedure. In addition, the solver guarantees the correctness of execution, similar to the interpreter used in program-aided LMs (\pallm{}).

We evaluate our approach on 8 datasets spanning 4 tasks, including arithmetic reasoning, logical reasoning, symbolic reasoning, and a regex synthesis task. Our \satlm{} consistently outperforms {\sc CoT} and \pallm{} across all datasets, usually by a large margin. On \gsmsys{}, \satlm{} outperforms \pallm{} by a 23\%; on \gsm{}, \satlm{} achieves 84.8\% with self-consistency decoding using few-shot prompting, equaling past work that uses the full training set and the same LLM~\citep{li2022advance,ni2023lever}. \satlm{} also sets a new SoTA on \lsat{}~\citep{arlsat}, \boardgame{}~\citep{boardgameqa}, and {\sc StructuredRegex}~\citep{structuredregex}.

Our analysis illustrates why the combination of SAT solver and declarative prompting is so effective.
We find (1) program-aided LMs often make planning errors (e.g., manipulating equations incorrectly), which can be remedied by the SAT solver. (2) Forcing LLMs to explicitly state a declarative specification can even improve vanilla CoT prompting. (3) Our \satlm{} approach can abstain from making uncertain predictions if it parses a problem into an unsatisfiable or ambiguous specification, giving it even higher accuracy in the selective prediction setting~\citep{selectivesetting}.

\section{Overview}

This work addresses the challenge of using LLMs to solve NL reasoning tasks. At a high level, an NL reasoning task is a natural language description of a collection of facts $\Phi$ (such as propositions or constraints) about some objects and a question $\query$ related to these objects. The goal of the reasoning task is to find an answer to $\query$ that can be deduced from the information provided in $\Phi$.

We conceptualize the general procedure for solving NL reasoning tasks in three steps: \emph{parsing}, \emph{planning}, and \emph{execution}. We are given natural language input $x_{\tt test} = (NL(\spec), NL(\query))$ which describes both $\spec$ and $\query$. Our first step is to parse this natural language into a predicted \emph{task specification} $(\hat{\spec}, \hat{\query})$, which is a \emph{formal} description of the facts and the query.

Given $(\hat{\spec},\hat{\query})$, the planning step then involves determining a sequence of reasoning steps $[r_1,\ldots,r_n]$ beginning with the task specification and ending with the answer to the question. Each step involves invoking a function (e.g., arithmetic operator or logical operator) that produces intermediate results which can be utilized in subsequent steps. A plan can be formulated by an LLM with {\sc CoT} prompting or by a symbolic solver as in our work here.
Finally, we execute the plan systematically with either a symbolic executor (our method) or an LLM, returning the output of the last step, $r_n$, as the answer.

Our solution approaches the problem using exactly these three steps. 

\paragraph{Parsing into declarative specification}
We prompt an LLM to generate a specification $s_{\tt test}$ for $x_{\tt test}$. Note that the translation from this description into the specification is not straightforward and cannot be done in a rule-based way for most tasks; Figure~\ref{fig:exs_qualitative} shows some particularly complex examples involving commonsense reasoning. The specification $s_{\tt test}$ is a sequence of interleaved NL statements and logical formulas (LF): $s_{\tt test}=[z_1,\ldots,z_n]$ and $z_i\in \Sigma_{NL}\cup \Sigma_{LF}$, where $\Sigma_{NL}$ and $\Sigma_{LF}$ denote the space of natural language and logical formulas, respectively.  We derive the formal specification $(\hat{\spec},\hat{\query})$ by taking all the $z_i$ in $\Sigma_{LF}$ from $s_{\tt test}$.
An example of the specification is presented on the right of Figure~\ref{fig:framework}.
Our specification is declarative since we do not explicitly generate the $r_i$ from the LLM at this stage. 

\paragraph{Planning and execution with a SAT solver} Given the predicted formal specification $(\hat{\spec}, \hat{\query})$, we wish to derive the final answer of the query $\hat{\query}$  from it. 
We say that a solution $a$ is correct if $\hat{\spec}$ logically entails $\hat{\query}=a$, denoted as $\hat{\spec} \models \hat{\query} = a$. The key insight behind our work is to offload \emph{both} the planning and execution steps to a SAT solver.
Specifically, we use a SAT solver to find a satisfying assignment for $a$ in the formula:
\footnotesize
\[
\forall V. \ (\hat{\spec} \Rightarrow \hat{\query} = a)\]
\normalsize
where $V$ denotes the set of all variables used in $\hat{\spec}$ and $\hat{\query} \in V$ is a variable that corresponds to the solution. Note that the only free variable in this formula is $a$; hence, the assignment to $a$ returned by the solver is the final answer to the reasoning problem.

The approach outlined above has two important strengths. First, because the SAT solver is \emph{sound} (i.e., any assignment it produces satisfies the formula), the solution is correct by construction. Thus, assuming that the parsing is correct and $\hat{\spec}$ and $\hat{\query}$ match $\spec$ and $\query$, we have a proof that the solution is indeed correct. Second, the planning step is done internally to the solver, and the chain of reasoning steps $[r_1, \ldots, r_n]$ can be obtained by asking the solver to produce a proof of the validity of the formula $\hat{\spec} \Rightarrow \hat{\query} = a^*$ where $a^*$ is the assignment produced by the SAT solver. All solvers we consider can produce such a proof of validity (e.g., in the form of a resolution refutation~\citep{resolution}).



\begin{figure}[t]
  \begin{center}
    \includegraphics[width=1.0\linewidth,trim=20 420 30 0,clip]{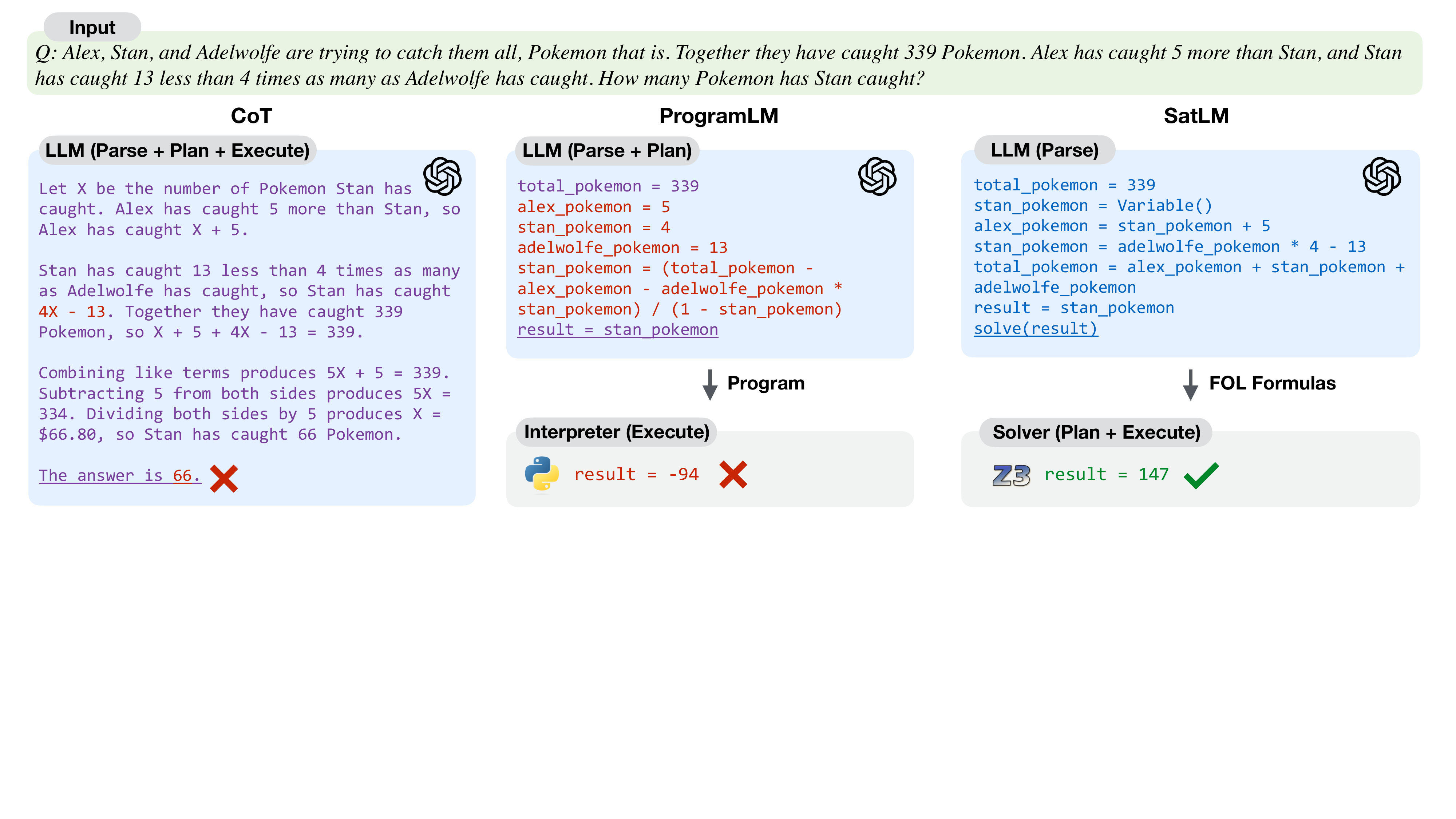}
  \end{center}
  \caption{Exemplar specifications for arithmetic reasoning problems generated by different approaches. \cotlm{} makes errors when parsing an equation; \pallm{} produces an incorrect reasoning chain (both errors are highlighted in \textcolor{red}{red}). By only using the LLMs to generate declarative specifications and relying on a solver to handle the reasoning, \satlm{} generates the correct answer.}
  \vspace{-0.5em}
  \label{fig:math_exs}
\end{figure}

\paragraph{Comparison with prior work} Prior approaches to NL-based reasoning with LLMs can also be framed in the parse-plan-execute framework proposed above. In particular, the chain-of-thought paradigm~\citep{scratch,chain} uses LLMs to perform each of the three steps. Program-aided language models~\citep{pal,progcot,faithfulcot} combine the parsing and planning steps to use an LLM to derive a program that corresponds to the plan.\footnote{This is true for ``faithful chain-of-thought'' as well \citep{faithfulcot}. This paper describes a breakdown of the process into ``translation'' and ``solving'' stages, where the translation step corresponds to both our parsing and planning stages. The solver used in that approach for tasks like CLUTRR does not do additional planning, but merely executes the steps outlined in CoT. In addition, their approach uses Python for execution, whereas ours uses SAT and Z3 as the unifying solving framework.} The final execution step is then performed by using the interpreter of the underlying programming language to derive the final answer. In contrast to these approaches, our work uses an LLM only to perform the parsing step, which is an easier problem for LLMs than planning.

We show a concrete example comparing \cotlm{} and \pallm{} with our approach in Figure~\ref{fig:math_exs}. \cotlm{} performs all three steps with the LLM. For instance, ``\emph{Alex has caught
$X + 5$}'' in the output corresponds to ``\emph{Alex has caught 5 more than Stan}'' in the NL input (parsing). Later, \cotlm{} decides how to solve for the variable $X$ with ``\emph{Combining like terms ...}'' (planning). At the same time, it also derives the equation ``$5X = 334$'' directly in its generation (execution). However, \cotlm{} incorrectly uses the same $X$ in the equation ``$X + 5$'' and ``$4X - 13$'', when it is supposed to be different. (Note that $4X-13$ would be correct if Stan and Adelwolfe's roles in the corresponding NL clause were reversed.) By allowing the LLM to focus only on translation, we find a lower incidence of this kind of error, in addition to eliminating planning errors. Notably, planning errors are \textbf{not} addressed by \pallm{}, which does not use programmatic manipulation at this stage. 
Different from \pallm{}, \satlm{} only parses the information provided in the input question, passes the parsed formulas to a solver for both planning and execution, and obtains the correct result.

\section{SAT-Aided Language Models using Declarative Prompting}




\subsection{Declarative Prompting}
\label{sec:sat_prompt}

We use few-shot prompting to generate the specification $s_{\tt test}$ for the test input $x_{\tt test}$. Specifically, we include few-shot demonstrations $(x_i,s_i)_{i=1}^k$ in the prompt, append test input $x_{\tt test}$ after the prompt, and let the LLM complete the specification for $x_{\tt test}$, i.e., $s_{\tt test} \sim p(x_{\tt test} \mid x_1,s_1,\ldots,x_k,s_k)$.

We show an example specification for a logical reasoning task in Figure~\ref{fig:framework}, and an example specification for an arithmetic reasoning task in Figure~\ref{fig:math_exs}. Observe that in both examples, our SAT formulas (i.e., the logical formulas of $[z_1,\ldots,z_n]$ in $\Sigma_{LF}$) are written as code following Python syntax, while the natural language in $\Sigma_{NL}$ is written using comment syntax. We found that including the language here as comments was useful to improve the fidelity of the translation. Our declarative prompts also use meaningful variable names and descriptive comments following the style of prompts in prior work~\citep{pal,faithfulcot}. Finally, we use Python rather than a specialized DSL to be more congruent with our models' pretraining data~\citep{instructgpt, codex}. See Appendix~\ref{app:detailed_specification} for more details on the SAT specification.

\subsection{Solving with a SAT Solver}
\label{sec:sat_solver}


\paragraph{SAT problem} A SAT problem is a triple $\satprob = (\spec, \theory, \query)$ where $\spec$ is a set of first-order logic formulas in some theory $\theory$\footnote{The theory defines the meaning of some of the symbols used in the formula. For example, in the theory of linear arithmetic, axioms of the theory give meaning to operators like addition, less than, etc.}
and $\query$ is the query of interest. We use ${\tt Variable}(\satprob)$ to denote the free variables in $\Phi$. $Q$ contains only variables in ${\tt Variable}(\satprob)$. An example SAT problem is $\satprob = (\{x+y=3, x-y=1\}, \theory_{E} \cup \theory_{\mathbb{Z}}, x-2)$, where $\theory_{E} \cup \theory_{\mathbb{Z}}$ indicates that only equality and linear arithmetic operations on integers are allowed in the formulas.

Many NL reasoning tasks in the literature can be formulated as SAT problems and solved using an off-the-shelf  solver.
For \textbf{arithmetic reasoning}, the SAT formulas $\Phi$ are equations encoding the relationships between variables, and $t$ specifies the target variable asked in the question (see Figure~\ref{fig:framework}). For \textbf{logical reasoning}, $\Phi$ encodes preconditions and $t$ specifies the target statement posed by the question.
We also show that symbolic reasoning, regex synthesis, and other problems involving reasoning over arrays or strings can be handled in this framework.

Unlike prior work such as Faithful CoT~\citep{faithfulcot} that uses task-specific formulations and task-specific solvers for different problem types, all the tasks in this paper are formulated as general SAT instances that can be solved by a single solver (as described later in this section).

\paragraph{Parsing NL to a SAT problem} Recall that we obtain a specification $s_{\tt test}$ from a test NL task $x_{\tt test}$. To derive the SAT problem $\satprob_{\tt test}= (\hat{\spec}_{\tt test},\mathcal{T}_{\tt test},\hat{\query}_{\tt test})$ from $s_{\tt test}$, we extract the constraints $\hat{\spec}_{\tt test}$ and the target expression $\hat{\query}_{\tt test}$ (marked by \ttsmall{solve} in our prompt) by taking all the $z_i$ in $\Sigma_{LF}$ of $s_{\tt test}$. We identify the theory $\mathcal{T}_{\tt test}$ by analyzing the formulas in $\hat{\spec}_{\tt test}$.

\paragraph{Solving the SAT problem}
Given the SAT problem $\satprob$, we invoke an automated theorem prover (such as the Z3 SMT solver~\citep{z3} used in our implementation)  to obtain a model  ${M}$ that maps each free variable $v \in {\tt Variable}(\satprob)$ to a concrete value under theory $\theory$. The final answer is obtained by substituting each free variable $v_i$ in $\hat{\query}$ with $M[v_i]$. For example, given the problem $ (\{x+y=3, x-y=1\}, \theory_{E} \cup \theory_{\mathbb{Z}}, x-2)$,  we ask the solver to find a solution to the constraint $x+y=3 \land x-y=1$ in the theory  $\theory_{E} \cup \theory_{\mathbb{Z}}$, which yields  $x = 2$ and $y=1$. Then, to obtain the final answer, we substitute $x$ by $2$ in the target expression $x-2$ to obtain the result  $2 - 2 = 0$.


\paragraph{Feedback signals from the solver}
Given a set of $\hat{\spec}$ specified in $\satprob$, the SAT solver will try to search for a satisfying assignment $M$ which satisfies all constraint formulas in $\hat{\spec}$.  If the solver succeeds in finding such an assignment within a certain time limit, it will use $M$ to evaluate the query $\hat{\query}$ and return the final result, otherwise it is a timeout. However, the solver may fail to find a solution for problematic $\satprob$ and provide feedback in one of the following types: (1) 
\textbf{\emph{error in execution}} ({\small ERROR}) caused by invalid formulas (e.g., syntax errors) or time-out;  (2) \textbf{\emph{unsatisfiable formulas}} ({\small UNSAT}), caused by conflicting formulas in the $\hat{\spec}$ (e.g. $\hat{\spec}=\{x=y+1,y=x+1\}$) (no feasible solution); (3) \textbf{\emph{ambiguous formulas}} ({\small AMBIG}), caused by the existence of multiple feasible solutions (e.g. $\hat{\spec}=\{x=y+1,x>0\}$). 
 Examples of SAT formulas leading to  {\small UNSAT} or  {\small AMBIG} can be found in Appendix~\ref{app:exs_sat_errors}.
 
Unlike the executor used in \pallm{} that can only detect errors in code execution, SAT solver can spot  {\small UNSAT} and  {\small AMBIG} in addition to  {\small ERROR}. We show this unique characteristic allows our \satlm{} to abstain from potentially incorrect predictions much more effectively compared to \pallm{} 
 in the selective prediction setting~\citep{selectivesetting} (Section~\ref{sec:selective_pred}).

\section{Experiments}
\subsection{Setup}
\label{sec:setup}
\paragraph{Tasks} Our work investigates 8 datasets covering 4 tasks, with a focus on arithmetic reasoning and logical reasoning tasks. We list all dataset statistics in Appendix~\ref{app:dataset_stats}. For arithmetic reasoning, we use \gsm{}~\citep{gsm8k}, \gsmsys{}, and \algebra{}~\citep{gsmsat}. \gsmsys{} is a special subset of \gsm{} containing examples that are paired with human-annotated solutions involving systems of equations (see Appendix~\ref{app:dataset_stats} for more details). For logical reasoning, we use \lsat{}~\citep{arlsat}, \boardgame{}~\citep{boardgameqa}, \clutrr{}~\citep{clutrr}, and \proofwriter{}~\citep{proofwriter}.
For \boardgame{}, we report the average performance on the three data splits (depth 1 to depth 3).

For \clutrr{}, we use exemplars requiring up to 3 intermediate steps but evaluate on test examples requiring up to 10 intermediate steps~\citep{clutrr}, following past work~\citep{faithfulcot}. For \proofwriter{}, we evaluate on the most challenging examples requiring depth-5 proofs~\citep{proofwriter}. For symbolic reasoning, we use Colored Object (\colorobj{}) from BIG-bench~\citep{bigbench} as an exemplar task. This task can be abstracted as finding elements in a list under certain constraints. We also evaluate on a regex synthesis dataset, \stregex{}~\citep{structuredregex}, which requires synthesizing a regex give NL description. We cast this task into synthesizing the surface form (i.e., a string) of the target regex, and use \satlm{} to parse NL description into constraints over the string. 


\paragraph{Baselines}
We compare \satlm{} against 3 baselines, including standard prompting (directly giving the answer), chain-of-thought prompting ({\sc CoT}), and executor-augmented LLMs (\pallm{}). We do not compare to zero-shot baselines such as zero-shot CoT, which generally underperform few-shot CoT by a large margin on the tasks we investigate~\citep{zerocot}.

For {\sc CoT} and \pallm{}, we leverage prompts of existing work~\citep{pal,faithfulcot,creswell2023selectioninference} whenever possible. For \satlm{}, we manually write prompts for the \textbf{same exemplar sets} used in {\sc CoT} and \pallm{} to ensure a fair comparison. We note that some settings, such as \pallm{} for \lsat{}, are not applicable. Please refer to Appendix~\ref{app:detailed_baselines} for more discussion of the setup, including details on the prompts we use. We also include example prompts for all the datasets in Appendix~\ref{app:detailed_prompts}.

\paragraph{Language Models \& Decoding}
We conduct our main experiments and analysis on \ttsmall{code-davinci-002}~\citep{codex}, a state-of-art LLM for code and code-adjacent tasks. We also include results on a less capable version \ttsmall{code-davinci-001} and an LLM specialized more to NL, \ttsmall{text-davinci-003}. We evaluate the performance with both greedy decoding and self-consistency decoding~\citep{selfcons}.
Due to the high computation cost, we use 5 samples for \lsat{}, \boardgame{}, and \proofwriter{}, which involve long prompts, and use 40 samples for all other datasets.

\begin{table}[t]
  \caption{Comparison of our approach (\satlm{}) against standard prompting (directly predicting the answer), {\sc CoT} and \pallm{}. Certain settings are not applicable (marked as $-$) as described in Appendix~\ref{app:detailed_baselines}. With greedy decoding, \satlm{} outperforms {\sc CoT} and \pallm{} on all datasets  by a substantial margin except for \gsm{}, where it is on par with \pallm{}. With self-consistency decoding, \satlm{} is consistently better than \pallm{}, giving SoTA accuracy on \lsat{} and \boardgame{}.}
  \vspace{0.5em}
  \label{tab:main}
  \footnotesize
  \renewcommand{\tabcolsep}{1.3mm}
  \centering
  \begin{tabular}{lcccccccccccc}
    \toprule
         & \gsmsys{} & \gsm{} & {\sc Alge} & & \lsat{} & {\sc Board} & \clutrr{} &{\sc Proof} & & \colorobj{} &  & {\sc Regex}\\
        \cmidrule{2-4}\cmidrule{6-9}\cmidrule{11-11}\cmidrule{13-13}
    & \multicolumn{12}{c}{\it \footnotesize code-davinci-002 (greedy decoding)}\vspace{0.05in}\\
    \sc Standard            & 21.0 & 22.2& 45.9 & & 22.0   & 44.6 & 41.2& 76.6  && 75.7&  &    $-$\\
    \sc CoT                 & 46.5 & 62.7& 53.6 && 23.5 & 60.7 & 40.8 &  80.1 && 86.3  & & $-$ \\
    \sc  \pallm{}   & 43.4 & \bf 72.7 & 52.3 && $-$ & $-$  & 58.9 &   83.7 && 95.1 &  & 39.1\\
     \sc SatLM            & \bf 69.4 & 71.8 &\bf 77.5 & & \bf 35.0 & \bf 79.4 & \bf 68.3 & \bf 99.7 && \bf 97.7 & & \bf 41.0\\
        \cmidrule{2-4}\cmidrule{6-9}\cmidrule{11-11}\cmidrule{13-13}
           & \multicolumn{12}{c}{\it \footnotesize  code-davinci-002 (self-consistency decoding)}\vspace{0.05in}\\
    \sc CoT                 & 56.1  & 77.3& 64.9 && 23.1  & 62.8 & 45.7 & 88.7 & & 90.6 & &  $-$\\
    \sc  \pallm{}   & 53.4& 82.4 & 57.7 & & $-$ & $-$ & 71.9 &  91.2 && 98.0  & & 56.5   \\
    \sc  SatLM   & \bf 80.9 & \bf 84.8 & \bf 90.9& & \bf 37.4 & \bf 80.7 & \bf 80.1&  \bf 99.7 && \bf 99.4  && \bf 59.7 \\
    \bottomrule
  \end{tabular}
\end{table}

\subsection{Main Results}
Table~\ref{tab:main} shows the performance of our approach compared to the baselines. In general, our {\sc Sat}-aided approach outperforms both {\sc CoT} and \pallm{} by a substantial margin except on \gsm{} with greedy decoding. We perform significance tests via bootstrap resampling, and all improvements of \satlm{} over \pallm{} are statistically significant ($p < 0.05$).

The first two columns show the performance on the \gsm{} dataset. \cotlm{} and \pallm{} achieve much worse performance on \gsmsys{} than on \gsm{}, indicating that \gsmsys{} is a challenging subset. On this subset, \satlm{} achieves 69.4\% and 80.9\% with greedy decoding and self-consistency decoding, surpassing both \pallm{} and {\sc CoT} 
 more than by 20\%. On the original \gsm{} dataset, the \satlm{} model has a slightly lower accuracy than \pallm{} with greedy decoding, but outperforms it with self-consistency decoding by 2.4\%; we provide detailed analysis accounting for the differences later in this section.
This self-consistency accuracy of 84.8\% even exceeds recent work that uses the full training set with \ttsmall{code-davinci-002} (82.3\% in {\sc DiVeRSe}~\citep{li2022advance}; 84.5\% in {\sc Lever}~\citep{ni2023lever}). On \algebra{}, a challenging dataset of math problems extracted from algebra textbooks, \satlm{} also outperforms \cotlm{} and \pallm{} by more than 20\%.
 
On \lsat{}, \clutrr{}, \proofwriter{}, and \colorobj{}, \satlm{} consistently achieves the best performance with either greedy decoding or self-consistency decoding. 
\satlm{} also sets the new SoTA on both \lsat{} and \boardgame{}, surpassing previous models that are trained on the full training set. Specifically, \satlm{} elevates the SoTA from 30.9\%~\citep{arlsat} to 37.4\% on \lsat{} and from 73.9\% ~\citep{boardgameqa}) to 80.7\% on \boardgame{}. See Appendix~\ref{app:boardgame_breakdown} for detailed performance breakdown on depth 1-3.

In the regex synthesis domain, with greedy decoding, directly translating natural language descriptions to regexes (\pallm{}) achieves 37.1\%, whereas using declarative prompting achieves 44.0\%. With self-consistency, we surpass the previous SoTA performance of 55.6\%~\citep{opsynth}.

\subsection{Impact of SAT Solver \& Declarative Prompting}
\label{sec:disentangle}
\begin{figure}
\begin{minipage}{0.48\textwidth}

  \begin{center}
    \includegraphics[width=\linewidth,trim=200 130 200 130,clip]{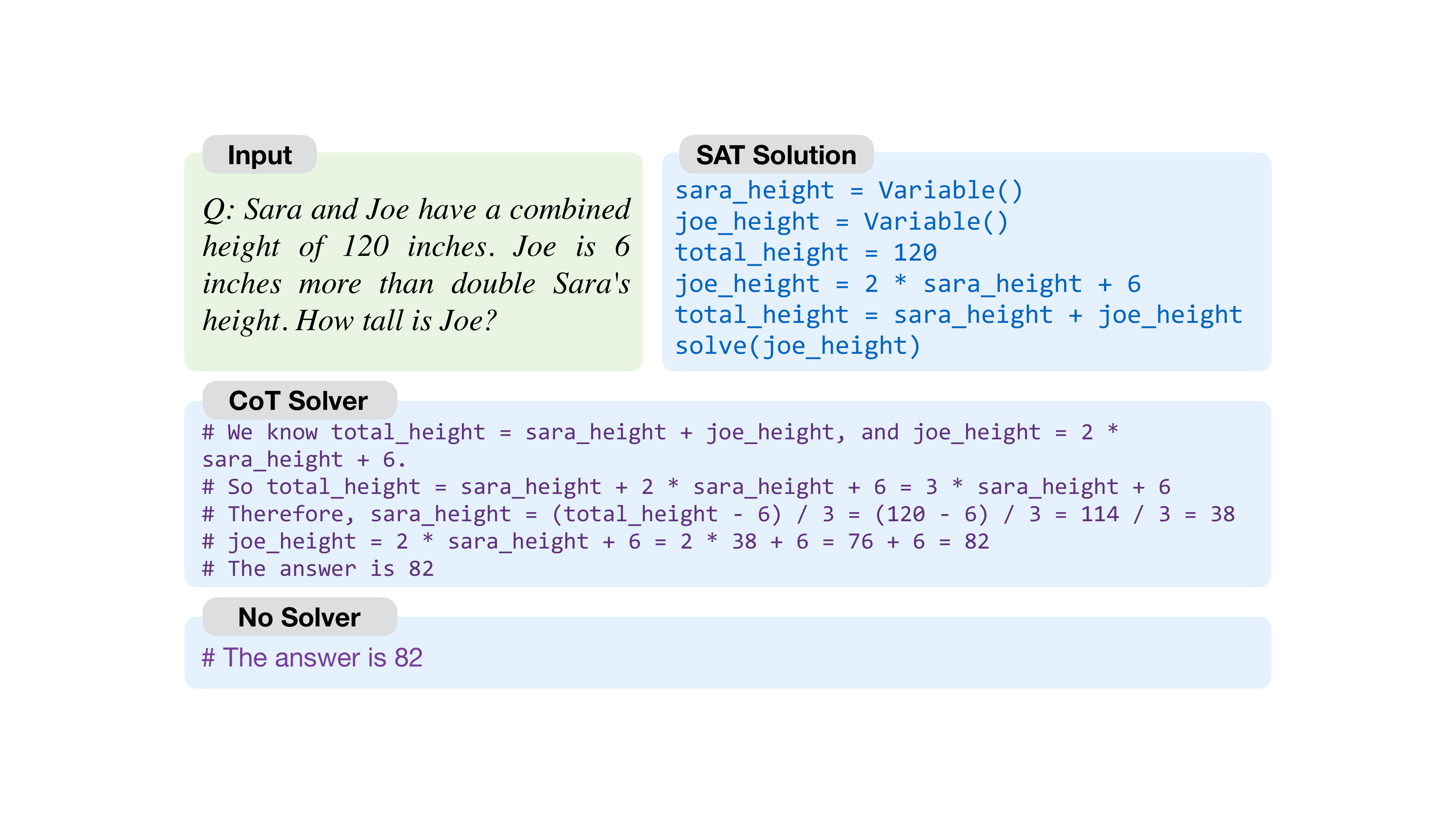}
  \end{center}
  \vspace{-.1in}
  \caption{A variant of our approach which replaces the SAT solver with a ``CoT solver'' that takes the SAT problem as input and solves it in natural language.}
  \label{fig:cot_solver}
  \end{minipage}
\hspace{0.01\textwidth}
  \begin{minipage}{0.48\textwidth}
\captionsetup{type=table}
  \caption{The performance of variants of our approach that use CoT Solver or No Solver. Using declarative prompting with CoT solver is more effective than imperative CoT prompting.}
  \label{tab:cot_solver}
  \vspace{0.1em}
  \small
  \centering
  \begin{tabular}{lccc}
    \toprule
         & \gsmsys{} & \gsm{} & \clutrr{} \\
        \cmidrule{2-4} 
    \sc Standard            & 21.0 & 22.2 & 41.2\\
    \sc CoT                 & 46.5 & 62.7 & 40.8 \\
    \sc PAL   & 43.4 &  72.8 & 58.9 \\
    \cmidrule{2-4}
     \sc Sat\textsubscript{SymSolver} &  69.4 & 71.7 &  68.3 \\
     \sc Sat\textsubscript{CotSolver} &  54.5 & 63.2 & 48.9 \\
     \sc Sat\textsubscript{NoSolver} & 26.6 & 23.7 &  40.7\\
    \bottomrule
  \end{tabular}
\end{minipage}
\end{figure}


We conduct analysis to isolate the effectiveness of the two key components, the SAT solver and declarative prompting. Specifically, we test a variant of our approach that still uses declarative prompting but then solves the equations in natural language with CoT rather than using the symbolic solver (see Figure~\ref{fig:cot_solver}). Essentially, the LLM itself carries out planning and execution. This experiment helps isolate the benefits of the solver, which will compute an answer without making any mistakes, from the benefits of the declarative formulation.  We also compare to prompting LLMs to directly give the answer ({\sc NoSolver}). 

\begin{wraptable}{r}{0.46\textwidth}
    \caption{Fraction of planning errors (incorrect reasoning chains) and execution errors (numeric errors) made by {\sc CoTSolver}. }
    \centering
    \footnotesize
    \begin{tabular}{lccc}
    \toprule
         & \gsmsys{} & \gsm{} & \clutrr{} \\
        \cmidrule{2-4} 
    \sc Plan Err  &  72.5 & 42.5 & 47.5 \\
    \sc Exec Err   & 27.5 & 57.5 & 52.5 \\
    \bottomrule
    \end{tabular}
    \label{tab:cotsolver_error}
\end{wraptable}

\paragraph{Impact of Symbolic Solver}  As shown in Table~\ref{tab:cot_solver}, completely ablating the solver and directly predicting the answer ({\sc Sat\textsubscript{NoSolver}}) only yields performance that is on par with {\sc Standard}.
Interestingly, {\sc Sat\textsubscript{CoTSolver}} can solve more SAT problems than {\sc NoSolver}. This partially reflects the effectiveness of CoT and partially reflects the fact that many dataset instances require relatively simple planning and execution, allowing pure forward reasoning to solve them. 
However, using a symbolic solver ({\sc Sat\textsubscript{SymSolver}}), which guarantees correct planning and execution, leads to further improvements. 

We manually analyzed 40 cases where the symbolic solver yields the correct answer but {\sc Sat\textsubscript{CoTSolver}} fails to solve them. We categorized the errors as planning errors, where the reasoning chains are incorrect, and execution errors, where the reasoning chains are correct but computations are incorrect (see Appendix~\ref{app:exs_cotsolver_errors} for examples). Table~\ref{tab:cotsolver_error} shows that most errors by {\sc Sat\textsubscript{CoTSolver}} are planning errors, 
especially on \gsmsys{} which requires solving complex system of equations. 

\begin{wraptable}{r}{0.42\textwidth}
    \caption{Log likelihood (unnormalized / normalized) of the generated sequences (with greedy decoding) of \pallm{} and \satlm{} on three datasets. Better log likelihood indicates higher LLM confidence in the parsing stage.}
    \centering
    \scriptsize
    \begin{tabular}{lccc}
    \toprule
         & \gsmsys{} & \gsm{} & \clutrr{} \\
        \cmidrule{2-4} 
    \sc PAL   & -9.5/-6.9\textsubscript{10\textsuperscript{-2}} & \bf -9.2/-6.0\textsubscript{10\textsuperscript{-2}} & -3.1/-8.5\textsubscript{10\textsuperscript{-3}} \\
    \sc SAT   & \bf -8.5/-5.9\textsubscript{10\textsuperscript{-2}} & -9.7/-6.2\textsubscript{10\textsuperscript{-2}} & \bf-2.0/-7.9\textsubscript{10\textsuperscript{-3}} \\
    \bottomrule
    \end{tabular}
    \label{tab:loglikelihood}
\end{wraptable} 

\paragraph{Impact of Declarative Prompting} Table~\ref{tab:cot_solver} also shows that decoupling parsing and planning/solving is still useful, even when not using a symbolic solver: {\sc Sat\textsubscript{CotSolver}} outperforms {\sc CoT} by 7.9\%, and 8.1\% on \gsmsys{} and \clutrr{}, respectively. We note that {\sc Sat\textsubscript{CotSolver}} can be viewed as a two-stage CoT prompting strategy, with a prompt showing that the first step is to formulate declaratively, then the next step is to solve. 

We hypothesize that parsing a question into declarative formulas is more straightforward than parsing it into an imperative solving procedure. To evaluate this hypothesis, we use log likelihood of the generated tokens to assess how straightforward the translation is, as higher log-likelihood typically indicates the outputs are more fluent to LLMs, a connection demonstrated in recent literature \citep{hila2022,optexpl}. We show both unnormalized (total) and normalized log likelihood in Table~\ref{tab:loglikelihood}. On \gsmsys{} and \clutrr{} where \satlm{} outperforms \pallm{}, its generated outputs are also associated with higher likelihood. 

\subsection{Advantages of {\sc Sat} in Selective Prediction}\label{sec:selective_pred}

A SAT solver may not always return an answer, particularly if there are parsing errors from the question. We show that this is an advantage of \satlm{}: these errors allow us to abstain from making likely incorrect predictions. 
Example outputs leading to different errors can be found in Appendix~\ref{app:exs_sat_errors}.

\begin{table}[t]
  \caption{Analysis of accuracy and execution status of \satlm{} and \pallm{}. We present the fraction of tasks solved correctly or incorrectly in \gsmsys{}, \gsm{}, and \clutrr{}, along with the breakdown of feedback from the solver. \satlm{} generally makes fewer predictions than \pallm{} (\textsc{Answered}), but more frequently makes correct predictions when it returns an answer (\textsc{Selective Acc}) \emph{and} gives a higher absolute number of correct predictions on \gsmsys{} and \clutrr{}.}
  \vspace{0.5em}
  \label{tab:exec_analysis}
  \scriptsize
  \renewcommand{\tabcolsep}{1.6mm}
  \centering
  \begin{tabular}{lccccccccc}
    \toprule
    & \multicolumn{2}{c}{\gsmsys{}} & & \multicolumn{2}{c}{\gsm{}} & & \multicolumn{2}{c}{\clutrr{}} \\
    & \pallm{} & \satlm{} & & \pallm{} & \satlm{} & & \pallm{} & \satlm{} \\
    \cmidrule{2-3}\cmidrule{5-6}\cmidrule{8-9}
    \sc Correct & 43.3 & 69.4 & & 72.7 & 71.8 & & 58.9 & 68.3\\
    \sc Incorrect & 52.5 & 20.6 & & 25.7 & 21.2 & & 21.0 & \phantom{0}7.7\\
     \cmidrule{1-1}
    \sc Error & \phantom{0}4.2 & \phantom{0}2.6 & & \phantom{0}1.6 & \phantom{0}2.1 & & 20.1 & \phantom{0}3.5\\
    \sc Unsat & $-$ & \phantom{0}2.4 & & $-$ & \phantom{0}1.5 & & $-$ & 15.5 \\
    \sc Ambig & $-$ & \phantom{0}5.0 & & $-$ & \phantom{0}3.4 & & $-$ & \phantom{0}5.0 \\
    \midrule
\sc Answered      & 95.8 & 90.0 & & 98.4 & 93.0 & &79.9 & 76.0\\
\sc Selective Acc & 45.2 & \bf 77.1 & & 73.8 & \bf 77.2& & 73.7 & \bf 89.9\\
    \bottomrule
  \end{tabular}
\end{table}

    


Table~\ref{tab:exec_analysis} shows the fraction of correct predictions and incorrect predictions when the program or SAT solver successfully returns an answer as well as the fraction of different types of feedback signals. We report the fraction of questions \emph{answered} as well as \emph{selective accuracy}, defined by the fraction of overall accuracy (\% of correct answers) normalized by coverage (\% of answered problems). \satlm{} makes fewer predictions on all three datasets compared to \pallm{}, as it can trigger both {\sc Unsat} and {\sc Ambig} errors. However, \satlm{}'s selective accuracy is consistently better than \pallm{}'s, especially on \gsmsys{} (77\% vs 45\%). As a result, \satlm{}'s overall performance is significantly better than \pallm{} on \gsmsys{} and \clutrr{}, even when making fewer predictions.

We note that on \gsm{}, \satlm{} has slightly lower coverage but higher selective accuracy compared to \pallm{}. This explains why \satlm{} lags behind \pallm{} with greedy decoding but outperforms \pallm{} with self-consistency decoding (Table~\ref{tab:main}). By drawing multiple samples, \satlm{} can increase its coverage and achieve higher accuracy than \pallm{} since its predictions are more accurate.\looseness=-1

\subsection{Analysis}

\begin{figure}[t]
  \begin{center}
    \includegraphics[width=0.9\linewidth,trim=0 585 0 0,clip]{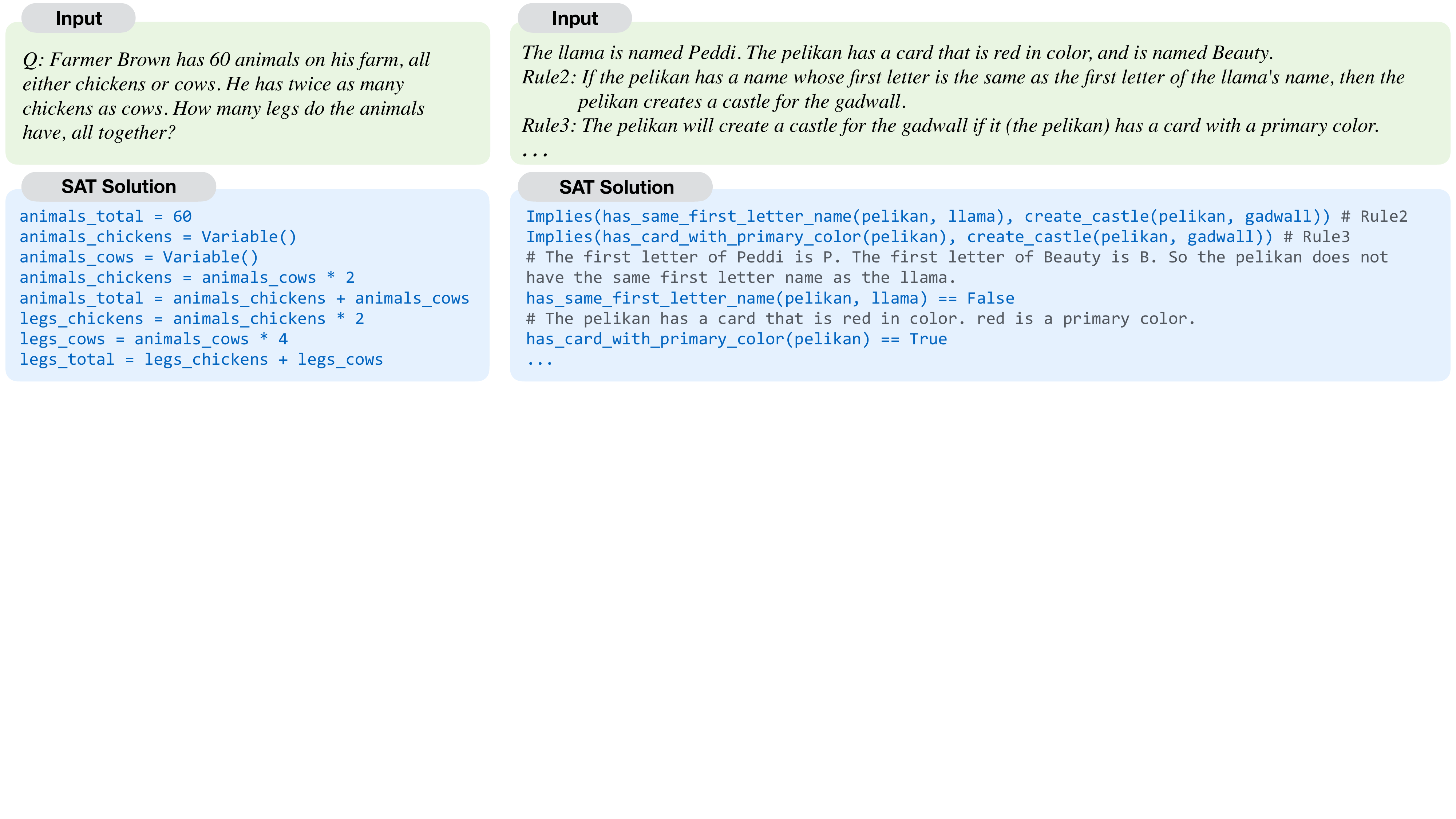}
  \end{center}
    \vspace{-0.5em}
  \caption{Examples outputs from \gsm{} (left) and \boardgame{} (right) show that LLMs can perform commonsense reasoning while parsing.}
  \label{fig:exs_qualitative}
\end{figure}

\paragraph{LLMs Can Perform Commonsense Reasoning While Parsing} There are many problems that do not state premises or constraints in a completely explicit way. Figure~\ref{fig:exs_qualitative}) shows two examples where commonsense inferences are required during parsing. For example, on the left, the model must recognize that \emph{animals} refers to the chickens and cows collectively. Similarly, knowing that red is a primary color is needed to successfully apply rules on 
\boardgame{} (right). We observe from the outputs in both cases that LLMs are capable of implicitly performing commonsense reasoning and produce correct logical formulas in the parsing step. As shown in Table~\ref{tab:main}, \satlm{} exhibits strong performance on \boardgame{}, a dataset which requires this implicit background knowledge.

\begin{wraptable}{r}{0.48\linewidth}
\vspace{-0.1in}
    \caption{Results on \ttsmall{gpt-3.5-turbo}, \ttsmall{text- davinci-003}, and \ttsmall{code-davinci-001}. The effectiveness of \satlm{} can generalize across LLMs. }
  \vspace{0.5em}
  \label{tab:other_llms}
  \scriptsize
  \renewcommand{\tabcolsep}{1.2mm}
  \centering
  \begin{tabular}{lcccccc}
    \toprule
         & \gsmsys{} & \gsm{} & & \lsat{} & \clutrr{} &{\sc Proof} \\
        \cmidrule{2-3}\cmidrule{5-7} 
        & \multicolumn{6}{c}{\it \footnotesize gpt-3.5-turbo (greedy decoding)}\vspace{0.05in}\\
        \cotlm{} & 44.8	& 74.4 &&  23.9	& 41.2 &	82.3\\
    \pallm{}   & 51.2	& \bf 77.9	&& 	$-$ & 45.9 &76.4\\
     \satlm{}    & \bf 63.4 & 76.4 && \bf 30.0 & \bf	50.6 & \bf 96.4 \\
        \cmidrule{2-3}\cmidrule{5-7} 
        & \multicolumn{6}{c}{\it \footnotesize text-davinci-003 (greedy decoding)}\vspace{0.05in}\\
    \cotlm{} & 42.8 & 62.5 && 21.7 & 34.5 & 83.5 \\
    \pallm{}   & 40.4 & \bf 71.7 && $-$  & 41.2 &   83.7 \\
     \satlm{}            & \bf 63.6 & 70.3 & & \bf 30.4 & \bf 58.2 & \bf 99.7 \\
        \cmidrule{2-3}\cmidrule{5-7} 
        & \multicolumn{6}{c}{\it \footnotesize code-davinci-001 (greedy decoding)}\vspace{0.05in}\\
    \pallm{}  & 15.5& \bf 35.6 & & $-$ & 22.2 &  63.8   \\
    \satlm{}   & \bf 16.5 & 34.2 & & 19.6 & \bf 30.2&  \bf 86.6  \\
    \bottomrule
 \end{tabular}
\vspace{-0.1in}
\end{wraptable}


\paragraph{Results Across Different Language Models}
In addition to the main LLM used in our work, \ttsmall{code-davinci-002}, we further test whether \satlm{} can generalize to other LLMs. We choose \ttsmall{gpt-3.5-turbo} (0613 version), \ttsmall{text-davinci-003}, and \ttsmall{code-davinci-001}. \ttsmall{gpt-3.5-turbo} is optimized for chat. \ttsmall{text-davinci-003} is an LLM pretrained on NL, and tuned to align with human feedback~\citep{instructgpt}. \ttsmall{code-davinci-001} is also an LLM pretrained on code, but less capable compared to \ttsmall{002}. As shown in Table~\ref{tab:other_llms}, \satlm{} is better than \pallm{} on the arithmetic reasoning and logical reasoning datasets except for \gsm{} across these three LLMs. The trend is congruent with the results on \ttsmall{code-davinci-002} (Table~\ref{tab:main}), which suggests the approach's general applicability across different LLMs, regardless of their varying capabilities.

\begin{wraptable}{r}{0.40\textwidth}
\vspace{-0.1in}
    \caption{The performance of \pallm{} and \satlm{} with varying exemplar sets. \satlm{} consistently outperforms \pallm{} on \gsmsys{} and \clutrr{}.}
    \vspace{0.5em}
    \centering
    \scriptsize
    \begin{tabular}{clccc}
    \toprule
        & & \gsmsys{} & \gsm{} & \clutrr{} \\
        \cmidrule{2-5} 
    \multirow{2}{*}{\STAB{\rotatebox[origin=c]{90}{Set1}}}
    & \sc Prog   & 43.4 & \bf 72.7 & 58.9 \\
    & \sc Sat   & \bf 69.4 & 71.8 & \bf 68.3 \\
    \cmidrule{2-5} 
    \multirow{2}{*}{\STAB{\rotatebox[origin=c]{90}{Set2}}}
    & \sc Prog   & 41.4 & \bf 72.5 & 59.0 \\
    & \sc Sat  & \bf 71.8 & 71.3 & \bf67.9 \\
    \cmidrule{2-5} 
    \multirow{2}{*}{\STAB{\rotatebox[origin=c]{90}{Set3}}}
    & \sc Prog   & 37.1 & \bf 70.3 & 57.2 \\
    & \sc Sat   & \bf 66.7 & 70.0 & \bf68.0 \\

    \bottomrule
    \end{tabular}

    \label{tab:var_exs}
\end{wraptable}

\paragraph{Sensitivity to Different Exemplar Sets}
We test whether the advantages of \satlm{} is sensitive to different sets of exemplars. We experiment with 3 sets of exemplars on \ttsmall{code-davinci-002}.
As shown in Table~\ref{tab:var_exs}, \satlm{} consistently outperforms \pallm{} by a large margin on \gsmsys{} and \clutrr{}, and achieves comparable performance on \gsm{}. The results suggest the effectiveness of our approach is insensitive to varying the choice of exemplars.


\section{Related Work}
Our work is built on top of few-shot prompting~\citep{gpt3}, which has proven effective on a wide range of tasks~\citep{wei2022emergent, promptsurvey, gem, recipe, flan, sanh2022multitask}. In particular, we focus on improving LLMs on reasoning tasks, which are challenging for language models even with recent developments~\citep{marcus2020next,garcez2023neurosymbolic}. Various techniques have been proposed for improving reasoning abilities~\citep{scratch,LeasttoMostPE, zerocot,khot2022decomposed, fu2022complexity,pinto,li2022explanations,faithfulcot}. They largely follow a chain-of-thought~\citep{chain} or scratchpad~\citep{scratch} paradigm. Among them, our work is the most related to the line of work that generates imperative programs to be executed by a symbolic executor, such as a Python interpreter~\citep{pal,progcot} or domain-specific executors~\citep{faithfulcot}. In this work, we propose a different paradigm that parses NL problems into declarative SAT problems and offloads the solving procedure to a SAT solver.

Previous work has also explored equipping LLMs with other tools, including search engines~\citep{yu2023generate,toolformer}, calculators~\citep{gsm8k,palm}, or other domain-specific special modules~\citep{toolformer,neuralsymlm}. A line of work focuses on using program-related tools such as program executors~\citep{poesia2022synchromesh}, program analysis tools~\citep{jigsaw}, and synthesis tools~\citep{marriage} to enhance the quality of the generated code. Our works further explores improving LLMs with SAT solvers.

Concurrent work explores the intersection of LLMs and planning, parsing planning problems into PDDL descriptions and leveraging a classical planner to produce the plan~\citep{liu2023llmplanner}. Our work differs in that we use the SAT formulation to solve general reasoning tasks, including arithmetic reasoning and logical reasoning, which cannot be specified in PDDL.

Also concurrently, \citet{gsmsat} combine LLMs and symbolic solvers for solving math problems. However, this work \emph{only} focus on arithmetic reasoning tasks and employs a math-specific symbolic solver (PySym). Our work takes a more general approach by formulating the problem within the scope of first-order logic and therefore is domain-agnostic. We also provide results of \satlm{} on the \algebra{} dataset collected by~\citet{gsmsat} in Appendix~\ref{app:discuss_concurrent}.

\section{Conclusion \& Limitations}

We have presented a framework for satisfiability-aided language models, casting a wide range of reasoning tasks into SAT problems under a unified formulation. We use an LLM to parse an NL query into a declarative specification and leverages a SAT solver to derive the final answer. Evaluation results on 8 datasets spanning 4 tasks across several LLMs demonstrate the effectiveness of our approach over program-aided language models. 

\paragraph{Limitations}
Our framework parses an NL problems into a set of declarative formulas. The NL description of some problems may already be more compatible with an imperative solving procedure, and our approach is likely to be less effective in these cases (e.g., \satlm{} slightly lags \pallm{} on \gsm{}). Future research can explore an integration or ensemble of these two prompting styles for more flexible reasoning.


\satlm{} heavily relies on the SAT solver and inherits some limitations of the SAT solver itself, such as computational cost when dealing with complex formulas involving quantifiers or nonlinear arithmetic. Moreover, SAT solvers can be limited by the expressiveness of the underlying theory, as not all theories can be easily encoded in first-order logic. Nevertheless, the wide range of tasks that we can instantiate our \satlm{} framework on shows its general applicability.

Our current approach parses a problem into a SAT specification, runs the solver, and returns the answer in a one-round fashion. One can imagine that unsatisfiable formulas or ambiguous formulas could be improved by re-prompting the model to improve the specification based on the exception signals, as explored in concurrent work for other problems~\citep{paul2023refiner,madaan2023self,chen2023teaching}. We believe this is an exciting direction for future work.

\section*{Acknowledgments}

Thanks to anonymous reviewers for their helpful feedback. This work was partially supported by National Science Foundation under Grants  No.2145280, No.1918889, No.1762299, No.2210831 and the NSF AI Institute for Foundations of Machine Learning (IFML). We would also like to thank authors of Pal~\citep{pal} and Faithful CoT~\citep{faithfulcot} for providing the prompts used in the baselines.

\bibliography{neurips_2023}
\bibliographystyle{neurips_2023}
\newpage
\appendix

\section{Detailed Statistics of Datasets}
\label{app:dataset_stats}

We show the statistics of all the datasets used in our paper in Table~\ref{tab:dataset_stats}.

For \clutrr{}, we follow the setting in {\sc FaithfulCoT}~\citep{faithfulcot}: we construct the prompt using exemplars requiring 2-3 reasoning steps and test whether the model can generalize to examples requiring up to 10 steps. We used the pre-processed test data consisting of 1,042 test examples from past work \citep{faithfulcot}. 

For \proofwriter{}, we use the closed world assumption setting, following past work~\citep{creswell2023selectioninference}. We construct our test set by randomly sampling a subset of 1,000 examples (out of 10,000) from the test split of depth-5 setting, the most challenging setting.

For \stregex{}, we merge the test and test-E split (see~\citet{structuredregex}) to form a test set consisting of 996 examples in total.

\begin{table}[h]
    \caption{Number of few-shot exemplars, number of test examples and license for the datasets used in our paper.}
    \centering
    \scriptsize
    \begin{tabular}{lccc}
    \toprule
        & \bf \# Shot &\bf \# Test & \bf License \\
        \cmidrule{2-4} 
 \gsm{}~\citep{gsm8k}   & 8 & 1,319 & MIT license\\
\gsmsys{} & 8 & 547 &  MIT license \\
\algebra{}~\citep{gsmsat} & 8 & 222 & Creative Commons Attribution Share Alike 4.0 \\
        \cmidrule{2-4} 
\lsat{}~\citep{arlsat} & 8 & 230 & MIT license\\
\boardgame{}~\citep{boardgameqa} & 5 & 3,000 & CC BY 4.0. \\
\clutrr{}~\citep{clutrr} & 8 & 1,042 & Attribution-NonCommercial 4.0\\
\proofwriter{}~\citep{proofwriter} & 4 & 1,000 & CC BY 4.0. \\
        \cmidrule{2-4} 
    \sc ColoredObject (Big-Bench) & 3 & 2,000 & Apache 2.0\\
    \cmidrule{2-4}
    {\sc StructuredRegex}~\citep{structuredregex} & 8 & 996 & MIT license\\
    \bottomrule
    \end{tabular}
    \label{tab:dataset_stats}
\end{table}

\paragraph{\gsmsys{} Dataset}

We construct \gsmsys{}, a special subset consisting of 547 examples extracted from \gsm{}. Specifically, we filter the entire \gsm{} dataset (train split and test split) to find examples whose human-annotated explanations involve a system of equations, using patterns like \textit{``let [letter] be''}, \textit{``assume [letter] be''} and \textit{``{[number][letter]}''}. We manually inspected 10\% of the examples and found 80\% of those samples did involve systems of equations in the explanation. We refer to this more challenging dataset as \gsmsys{}.

\section{Details of the Prompts}
\label{app:detailed_baselines}

In general, we leverage {\sc CoT} prompts and \pallm{} prompts from existing work whenever available, and manually write \satlm{} prompts for the \textbf{same exemplar sets}. Prompt examples for all datasets can be found in Appendix~\ref{app:detailed_prompts}.


For \textbf{\gsm{} and \gsmsys{}}, we adapt the original {\sc CoT} prompt and \pallm{} prompt used in program-aided language models~\citep{pal}. Specifically, we replace one random exemplar in the original prompt with another exemplar sampled from \gsmsys{}. This is to improve the performance of {\sc CoT} and \pallm{} on \gsmsys{}, as the original exemplar set achieves suboptimal performance for \gsmsys{}. Our adapted {\sc CoT} and \pallm{} prompts achieve better performance compared to the original ones on both \gsm{}
 and \gsmsys{} (see Appendix~\ref{app:original_gsm_exs} for details).

 For \textbf{\lsat{}}, we randomly sample 8 exemplars and write prompts for {\sc CoT} and \satlm{}. We note that \lsat{} is a particularly challenging task: we tried 3 CoT prompts written by 3 different authors of our paper, which all led to around 20\% accuracy. Similar results are reported in other work~\citep{Liang2022HolisticEO,ribeiro2023street}. In addition, we only report \cotlm{} results, leaving out \pallm{}. This decision is due to the fact that \pallm{} uses Python as its program interpreter. While Python is a general-purpose programming language, it does not provide native support for formal logic reasoning, including essential components like logical inference rules and manipulation of logical formulas. Solving problems from \lsat{} requires strategies like proof by contradiction (see Appendix~\ref{app:detailed_prompts} for a detailed example), which we see no way to represent in the \pallm{} framework and is not addressed in prior work. 

\textbf{\boardgame{}} contains problems requiring 1-3 steps of reasoning. We sample 5 exemplars from the training set of depth 1 and depth 2 to construct the prompts for evaluation on the test sets of depth 1 and depth 2, respectively. We used the 5 exemplars of depth 2 to construct the prompt for test set of depth 3, as using exemplars of depth 3 would lead to prompts that exceed the context window size of our LLMs. Similarly, we only report \cotlm{} results as the baselines, leaving out \pallm{} for \boardgame{}. We use the proofs provided by the authors to construct the \cotlm{} prompts and manually annotate the SAT specifications to construct the \satlm{} prompts.

 For \textbf{\clutrr{}}, we use the {\sc CoT} prompt and \pallm{} prompt provided in {\sc FaithfulCoT}~\citep{faithfulcot}. For \textbf{\textsc{ProofWriter}}, we use the {\sc CoT} prompt from {\sc Selection-Inference}~\citep{creswell2023selectioninference}, and adapt it to form the \pallm{} prompt. We use the {\sc CoT} prompt and \pallm{} from {\sc Pal}~\citep{pal} for \textbf{\textsc{Colored Object}}.
 
The task of \textbf{\textsc{StructuredRegex}}, a regex synthesis dataset, is to parse natural language descriptions to regexes. This is not a typical reasoning dataset, and there is no {\sc CoT} prompt for this dataset. We randomly sample 8 exemplars and annotate the prompt for \pallm{} and \satlm{}. In this setting, \pallm{} directly translates NL descriptions into regexes (which are essentially programs), whereas \satlm{} parses an NL description into a set of constraints over the surface form of the regex.
Note that this dataset provides \emph{multimodal} specifications of regexes, featuring both NL descriptions and examples. The I/O examples can be used to reject synthesized regexes if they do not accept or reject the correct examples.
When we report results for self-consistency inference, we follow past work~\citep{opsynth} and filter out incorrect outputs using the I/O examples provided in the dataset~\citep{structuredregex}. This setting therefore checks consistency with something other than the model itself, but uses a similar computation budget as self-consistency, so we group it with those results.


\section{Performance of Original \textsc{CoT} and \pallm{} Prompts on Arithmetic Reasoning Datasets}
\label{app:original_gsm_exs}

\begin{table}[h]
    \caption{Performance of different approaches using our adapted exemplar set and the original exemplar set used in {\sc CoT} and {\sc Pal}.}
    \centering
    \footnotesize
    \begin{tabular}{lccccc}
    \toprule
        & \multicolumn{2}{c}{\sc Adapted (Ours)} &  & \multicolumn{2}{c}{
        \sc Original} \\
        & \gsmsys{} & \gsm{} &  & \gsmsys{} & \gsm{}\\
        \cmidrule{2-3} \cmidrule{5-6}
    \sc CoT  & 46.5 & 62.7 &  & 35.7 & 62.4 \\
    \pallm{} & 43.4  & \bf 72.7 & & 36.1& \bf 71.7 \\
    \satlm{} & \bf 69.4 & 71.8 & & \bf 66.7& 70.9 \\
    \bottomrule
    \end{tabular}
    \label{tab:official_cot}
\end{table}

Recall that we construct our arithmetic reasoning prompt used in Table~\ref{tab:main} by replacing one random exemplar in the original prompt used in {\sc CoT} and \pallm{} with an random example from \gsmsys{}. We show the performance of {\sc CoT}, \pallm{}, and our \satlm{} in Table~\ref{tab:official_cot} using our adapted exemplar set and original exemplar set in Table~\ref{tab:official_cot}.

Our adaptation significantly improves the performance of {\sc CoT} and \pallm{} on \gsmsys{}, and slightly improves the performance on \gsm{}. Furthermore, we still see that \satlm{} outperforms both {\sc CoT} and \pallm{} by a large margin on \gsm{}, using either our adapted set or the original set.

\section{Extended Discussion on Concurrent Work}
\label{app:discuss_concurrent}

\begin{table}[h]
    \caption{Performance of different approaches on \algebra{}.}
    \centering
    \footnotesize
    \begin{tabular}{lcc}
    \toprule
        & \algebra{} & \gsm{}\\
        \cmidrule{2-3}
    \sc CoT  & 53.6 & 62.4  \\
    \pallm{} & 52.3  &  72.7 \\
    \satlm{} (Ours) & 77.5 &  71.8 \\
    {\sc MathSym} \citep{gsmsat} & 76.3 & 69.4 \\
    \bottomrule
    \end{tabular}
    \label{tab:disc_concurrent}
\end{table}

Similar to our work, \cite{gsmsat} proposes to solve arithmetic reasoning problems by parsing the problem into a set of variables and equations and using an external solver to derive the final answer. While their formalization is restricted to arithmetic problems, we use SAT problems encoded with first-order logical formulas, which unify a wide range of reasoning tasks.

In addition, we also evaluate our approach on the \algebra{} dataset in \cite{gsmsat}, which consists of 222 examples from Algebra textbooks. We note that the results between ours and {\sc MathSym} are not directly comparable, as {\sc MathSym} picks a different exemplar set. As shown in Table~\ref{tab:disc_concurrent}, \algebra{} is more challenging than \gsm{}, and \satlm{} outperforms \pallm{} and {\sc CoT} by more than 20\%.

\section{Detailed Performance on the \boardgame{} Dataset}
\label{app:boardgame_breakdown}

\begin{table}[h]
    \centering
        \caption{Detailed performance on the \boardgame{} dataset.}
                \vspace{0.5em}
        \label{tab:boardgame_detail}
    \begin{tabular}{lccccc}

    \toprule
         & \sc depth 1 & \sc depth 2 & \sc depth 3 & & \sc Aggregated \\
        \cmidrule{2-4}\cmidrule{6-6}
    & \multicolumn{5}{c}{\it \footnotesize code-davinci-002 (greedy decoding)}\vspace{0.05in}\\
    \sc Standard & 52.5	& 42.8& 38.5 &&	44.6\\
        \cotlm{} & 64.7	& 60.8&56.5	& &60.1 \\
        \satlm{} & \bf 87.6	& \bf 81.7 & \bf 69.0 && \bf 79.4\\ 
                \cmidrule{2-4}\cmidrule{6-6}
    & \multicolumn{5}{c}{\it \footnotesize code-davinci-002 (self consistency decoding)}\vspace{0.05in}\\
            \cotlm{} & 65.9&	63.4&	59.0&&62.8 \\
        \satlm{} & \bf 88.0 & \bf 84.2 & \bf 70.1 && \bf 80.8 \\  
        \bottomrule
    \end{tabular}
    \label{tab:my_label}
\end{table}

Table~\ref{tab:boardgame_detail} shows the performance breakdown on depths 1-3 of the \boardgame{} dataset. \satlm{} outperforms \cotlm{} by a substantial margin across all depths. The performance of all approaches decreases as the depth increases.

\section{Details of the SAT Specification}
\label{app:detailed_specification}
To better utilize the parametric knowledge that LLMs have acquired from pretraining on vast amount of code data, our work uses a specification that largely follows and simplifies the syntax for specifying constraints used in z3py.\footnote{\url{https://z3prover.github.io/api/html/namespacez3py.html}}

\begin{figure}[h]
    \footnotesize
    \centering
    \begin{tabularx}{\linewidth}{X}
    \toprule
        \multicolumn{1}{c}{\bf Example SAT Specification} \\
         \midrule
\tt\small x = Variable() \# declare a variable \\
\tt\small People = [Alice, Bob] \# declare enum set \\
\tt\small Cities = [Austin, Boston] \# declare enum set \\
\tt\small Food = [Apple, Banana] \# declare enum set \\
\\
\tt\small visit = Function(People, Cities) \# declare function \\
\tt\small eats = Function(People, Food) \# declare function \\
\\
\tt\small  visit(Alice) != visit(Bob)  \# logic \\
\tt\small
\tt\small ForAll(x: People, Implies(visit(x) == Austin, eats(x) == Banana)) \# quantifier \\
    \bottomrule
    \end{tabularx}
    \caption{ Example of our SAT specification. The syntax is largely the same as that for specifying constraints in z3py. }
    \label{fig:exs_sat_specification}
\end{figure}

We give an example specification in Figure~\ref{fig:exs_sat_specification} demonstrating the synax for different types of statements. See Figure~\ref{fig:framework}, Figure~\ref{fig:math_exs}, and Appendix~\ref{app:detailed_prompts} for more examples. These formulas are close to the actual python code formulas used by z3py but are slightly modified to be more amenable to prompting. As a result, we use a postprocessing step to form the actual Z3 input. We implemented a simple parser that transforms these formulas into actual specifications used by z3py via string transformation (using regexes). For example, we transform [\ttsmall{ForAll(x: People, Implies(visit(x, Austin), eats(x, Banana)))}] into [\ttsmall{x = Variable(People) ForAll([x], Implies(visit(x) == Austin, eats(x) == Banana))}], which is the actual z3py code.
The transformed code can be executed using z3py to obtain the final answer.

\newpage

\section{Examples of Error Cases of \satlm{}}
\label{app:exs_sat_errors}

\subsection{Examples of Unsatisfiable and Ambiguous Formulas}
We show examples of formulas that are unsatisfiable or have ambiguous answers in Figure~\ref{fig:usat_exs_gsm}, Figure~\ref{fig:ambig_exs_gsm}, and Figure~\ref{fig:unsat_ambig_clutrr}.

\begin{figure}[h]
    \footnotesize
    \centering
    \begin{tabularx}{\linewidth}{X}
    \toprule
        \multicolumn{1}{c}{\bf \textcolor{red}{UNSAT} Formulas: \gsm{}} \\
         \midrule
    Q: If a rectangle has a width of 42 inches and an area of 1638, how many rectangles of the same size would reach a length of 390 inches? \\
    \\
  \tt  \textcolor{red}{rectangle\_width = 42}\\
  \tt \textcolor{red}{rectangle\_area = 1638} \\
  \tt  \textcolor{red}{rectangle\_length = 390} \\
  \tt  \textcolor{red}{rectangle\_length = rectangle\_area / rectangle\_width} \\
   \tt rectangles\_needed = rectangle\_length / rectangle\_width \\
  \tt  result = rectangles\_needed \\
   \tt solve(result) \\
    \midrule
Q: Jason's dog has a tail that's half the length of its body, and a head that's 1/6 the length of its body. If the dog is 30 inches long overall, how long is its tail? \\
\\
  \tt  dog\_length = 30 \\
 \tt   tail\_length = Variable() \\
  \tt  head\_length = Variable() \\
  \tt  \textcolor{red}{dog\_length = tail\_length + head\_length} \\
  \tt  \textcolor{red}{tail\_length = dog\_length * 1 / 2} \\
   \tt \textcolor{red}{head\_length = dog\_length * 1 / 6} \\
   \tt result = tail\_length\\
   \tt solver(result) \\
    \bottomrule
    \end{tabularx}
    \caption{ Examples of UNSAT formulas from \gsm{}. Both examples are generated by \ttsmall{code-davinci-002}. UNSAT occurs when LLMs misunderstand the problem and output conflicting constraints (highlighted in \textcolor{red}{red}).}
    \label{fig:usat_exs_gsm}
\end{figure}

\begin{figure}[h]
    \footnotesize
    \centering
    \begin{tabularx}{\linewidth}{X}
    \toprule
        \multicolumn{1}{c}{\bf \textcolor{red}{AMBIG} Formulas: \gsm{}} \\
         \midrule

    Q: \textcolor{red}{The sum of the three numbers is 500.} If the first number is 200, and the value of the second number is twice the value of the third number, find the value of the third number. \\
    \\
   \tt num1 = 200 \\
    \tt    num2 = Variable() \\
    \tt    num3 = Variable()\\
     \tt   num2 = num3 * 2 \\
      \tt  total = num1 + num2 + num3\\
      \tt  result = num3 \\
      \tt  solve(result) \\

    \midrule
Q: \textcolor{red}{The total for the sum and product of Elvie's age and Arielle's age are 131. } If Elvie's age is 10, how old is Arielle?
\\
\\
  \tt  elvie\_age = 10\\
   \tt arielle\_age = Variable()\\
   \tt sum\_age = elvie\_age + arielle\_age\\
 \tt   product\_age = elvie\_age * arielle\_age\\
  \tt  total\_age = sum\_age + product\_age\\
  \tt  result = arielle\_age\\
   \tt solve(result)\\
    \bottomrule
    \end{tabularx}
    \caption{ Examples of AMBIG solutions from \gsm{}. Both examples are generated by \ttsmall{code-davinci-002}. The ambiguous  formulas are under-constrained due to failure in encoding certain constraints specified in the problem (highlighted in \textcolor{red}{red}), resulting in multiple possible answers.}
    \label{fig:ambig_exs_gsm}
\end{figure}

\newpage

\begin{figure}[h]
    \footnotesize
    \centering
    \begin{tabularx}{\linewidth}{X}
    \toprule
        \multicolumn{1}{c}{\bf \textcolor{red}{UNSAT} Formulas: \clutrr{}} \\
         \midrule
        Arlene and her husband Jeff went on a cruise. They had a wonderful time. Stephanie s father Jason loves his little princess even though she gets into a lot of trouble at school. Gloria's mother Ruth and brother Jeff were working in the kitchen, preparing soup. Stephanie, Ruth's daughter, was working outside in the garden. \\
Q: How is [Jason] related to [Arlene]?\\
\\
	\# [Arlene] and her husband [Jeff] went on a cruise. They had a wonderful time. \\
   \tt relation(Arlene, Jeff) = (wife, husband)\\
    \# [Stephanie]'s father [Jason] loves his little princess even though she gets into a lot of trouble at school.\\
  \tt  relation(Stephanie, Jason) = (daughter, father)\\
    \# [Gloria]'s mother [Ruth] and brother [Jeff] were working in the kitchen, preparing soup.\\
   \tt relation(Gloria, Ruth) = (daughter, mother)\\
    \tt \textcolor{red}{relation(Gloria, Jeff) = (daughter, brother)}\\
    \# [Stephanie], [Ruth]'s daughter, was working outside in the garden.\\
  \tt  relation(Stephanie, Ruth) = (daughter, mother)\\
    \# How is [Jason] related to [Arlene]?\\
   \tt solve(relation(Jason, Arlene))\\
    \midrule
    \multicolumn{1}{c}{\bf\textcolor{red}{AMBIG} Formulas: \clutrr{} } \\
    \midrule
    Kirk loves talking to his grandfather Stanley on the phone. Paul's brother, Kirk, recently made it into town. Paul and his son Michael went to look at cars. Paul ended up buying the Mustang. \\
Q: How is [Stanley] related to [Michael]? \\
\\
   \# [Kirk] loves talking to his grandfather [Stanley] on the phone. \\
   \tt relation(Kirk, Stanley) = (grandson, grandfather) \\
    \# [Paul]s brother, [Kirk], recently made it into town. \\
   \tt relation(Paul, Kirk) = (brother, brother) \\
    \# [Paul] and \textcolor{red}{his son [Michael]}  went to look at cars. \\
   \tt \textcolor{red}{relation(Paul, Michael) = (father, son)} \tt \textcolor{blue}{\it note: Michael is Kirk's son, not Paul's son} \\
    \# How is [Stanley] related to [Michael]? \\
   \tt solve(relation(Stanley, Michael)) \\
    \bottomrule
    \end{tabularx}
    \caption{An example of UNSAT formulas (top) and an example of AMBIG formulas (bottom) from \clutrr{}. Both happen when the LLM misparses a constraint (highlighted in \textcolor{red}{red}). In particular, ambiguous formulas usually occur when the language description itself is highly ambiguous (see the note in the ambiguous formulas for an example). }
    \label{fig:unsat_ambig_clutrr}
\end{figure}

\newpage
\subsection{Examples of Incorrect Answers}

We show examples where the SAT solver successfully returns an answer, but the answer is incorrect in Figure~\ref{fig:incorrect_answers}.

\begin{figure}[h]
    \footnotesize
    \centering
    \begin{tabularx}{\linewidth}{X}
    \toprule
        \multicolumn{1}{c}{\bf Examples of Incorrect Answers: \gsm{}} \\
         \midrule
         Q: Martin rings the small bell 4 times more than 1/3 as often as the big bell. If he rings both of them a combined total of 52 times, how many times does he ring the big bell? \\
\\
  \tt  small\_bell\_rings = Variable() \\
   \tt big\_bell\_rings = Variable() \\
   \tt total\_bell\_rings = 52 \\
   \tt \textcolor{red}{small\_bell\_rings = big\_bell\_rings * 1 / 3 * 4 }\\
   \tt total\_bell\_rings = small\_bell\_rings + big\_bell\_rings \\
    \tt result = big\_bell\_rings \\
    \tt solve(result) \\
    \midrule
        Q: Jonathan was sad to learn he needed 2 more toys than he already had to have enough toys to make a sorted play set five times larger than James's play set, which had 80 toys. How many toys does Jonathan currently have? \\
        \\
  \tt  james\_toys = 80 \\
   \tt jonathan\_toys = Variable() \\
  \tt  \textcolor{red}{jonathan\_toys\_needed = james\_toys * 5 + 2} \\
   \tt result = jonathan\_toys\_needed - 2 \\
  \tt  solve(result) \\
    \midrule
     \multicolumn{1}{c}{\bf Examples of Incorrect Answers: \clutrr{}} \\
    \midrule

Donald and Russell are brothers. Jason is father of their father.\\
Q: How is [Russell] related to [Jason]?  \\
\\
	\# [Donald] and [Russell] are brothers.\\
  \tt  relation(Donald, Russell) = (brother, brother)\\
    \# [Jason] is father of their father \\
 \tt   \textcolor{red}{relation(Jason, Donald) = (father, son) }\\
  \tt  \textcolor{red}{ relation(Jason, Russell) = (father, son)} \\
    \# How is [Russell] related to [Jason]? \\
  \tt  solve(relation(Russell, Jason))\\
    \bottomrule
    \end{tabularx}
    \caption{Two examples of incorrect answers from \gsm{} (top) and an example of incorrect answer from \clutrr{} (bottom). This occurs when LLMs output incorrect constraints (highlighted in \textcolor{red}{red}) but the incorrect formulas can still derive a non-ambiguous answer. }
    \label{fig:incorrect_answers}
\end{figure}

\newpage
\section{Examples of Execution Errors and Planning Errors Made by {\sc CoTSolver}}
\label{app:exs_cotsolver_errors}

We show examples of execution errors and planning errors produced by {\sc CoTSolver} in Figure~\ref{fig:gsm_cotsolver_err_exec}, Figure~\ref{fig:gsm_cotsolver_err_plan},  and Figure~\ref{fig:clutrr_cotsolver_err}.

As shown in Figure~\ref{fig:gsm_cotsolver_err_exec}, LLMs can make errors when executing arithmetic operators, resulting in execution errors. While introducing an interpreter (\pallm{}) can help fix such execution errors, it is not effective for handling planning errors in Figure~\ref{fig:gsm_cotsolver_err_plan}, which involves manipulating symbolic variables to determine a chain of arithmetic equations leading to the final answer.

We note that for \clutrr{}, the atomic function is applying a set of  transitivity rules that infer the relation between \ttsmall{A} and \ttsmall{C} based on the relation between \ttsmall{A} and \ttsmall{B} and the relation between  \ttsmall{B} and \ttsmall{C} (e.g., \ttsmall{A's son is B, B's daughter is C. So A's granddaughter is C).} So an execution error in \clutrr{} refers to incorrect outputs when applying transitivity rules (see the top section of Figure~\ref{fig:clutrr_cotsolver_err}). 
The set of transitivity rules we use are provided in \cite{faithfulcot,zhang2022improved}.

\begin{figure}[h]
\footnotesize
    \centering
    \begin{tabularx}{\linewidth}{X}
    \toprule
       \multicolumn{1}{c}{\bf \textcolor{red}{Execution Errors} of {\sc CoTSolver}: \gsm{}}  \\
         \midrule
    Q: Carly is thinking about buying a wallet that costs $22$ and a purse that costs $3$ less than four times the cost of the wallet. What is the combined cost of both items? \\
    \\
 \tt  wallet\_cost = 22 \\
 \tt   purse\_cost = Variable() \\
  \tt  purse\_cost = wallet\_cost * 4 - 3 \\
  \tt  total\_cost = wallet\_cost + purse\_cost \\
  \tt  result = total\_cost \\
  \tt  \# We know purse\_cost = wallet\_cost * 4 - 3 \\ 
  \tt  \# So total\_cost = wallet\_cost + purse\_cost = 22 + (22 * 4 - 3) = 22 + \textcolor{red}{87  = 109} \\
  \tt  \# result = total\_cost = \textcolor{red}{109} \\
   \tt \# The answer is 109 \\
    \midrule
       Q: A person borrowed an amount of money for a year at an interest rate of 12\%. If the total interest is \$1500, what was the original borrowed amount? \\
       \\
  \tt  interest\_rate = 12\\
  \tt  interest\_total = 1500\\
   \tt borrowed\_amount = Variable()\\
  \tt  interest\_total = borrowed\_amount * interest\_rate / 100
   \tt result = borrowed\_amount\\
   \tt \# We know interest\_total = borrowed\_amount * interest\_rate / 100\\
  \tt  \# So borrowed\_amount = interest\_total / interest\_rate * 100 = 1500 / 12 * 100 = \textcolor{red}{12500 / 12 = 1041.67}\\
  \tt \# result = borrowed\_amount = 1041.67 \\
  \tt  \# The answer is 1041.67\\
\bottomrule
    \end{tabularx}
    \caption{Examples of execution errors made by {\sc CoTSolver} on arithmetic reasoning tasks. Both examples are generated by \ttsmall{code-davinci-002}, which makes mistakes when calculating the final value of equations.}
    \label{fig:gsm_cotsolver_err_exec}
\end{figure}

\begin{figure}[h]
    \footnotesize
    \centering
    \begin{tabularx}{\linewidth}{X}
    \toprule
         \multicolumn{1}{c}{\bf \textcolor{red}{Planning Errors} of {\sc CoTSolver}: \gsm{}} \\
         \midrule
Q: Griffin had 24 french fries, but Kyle took 5 of them. Billy took twice as many as Kyle. Ginger gave Griffin a handful of her fries, and then Colby took from Griffin 3 less than the number of fries that Kyle had taken. If in the end Griffin had 27 fries, how many fries did Ginger give Griffin? \\
\\
 \tt   fries\_initial = 24 \\
  \tt  kyle\_took = 5 \\
  \tt  billy\_took = kyle\_took * 2 \\
 \tt   ginger\_gave = Variable() \\
  \tt  colby\_took = kyle\_took - 3 \\
  \tt  fries\_after = 27 \\
  \tt  fries\_after = fries\_initial - kyle\_took - billy\_took + ginger\_gave - colby\_took \\
  \tt  result = ginger\_gave \\
  \tt  \# Solve the value of result \\
  \tt  \# We know fries\_after = fries\_initial - kyle\_took - billy\_took + ginger\_gave - colby\_took  \\
  \tt  \# So ginger\_gave = fries\_after - fries\_initial + kyle\_took + billy\_took - \textcolor{red}{colby\_took} = 27 - 24 + 5 + 10 - 2 = 16 \\
   \tt \# result = ginger\_gave = 16 \\
   \tt \# The answer is 16\\
\midrule
    Q: The combined age of Peter, Paul and Jean is 100 years old. Find the age of Peter knowing that Paul is 10 years older than John and that Peter’s age is equal to the sum of Paul and John's age. \\
    \\
  \tt  john\_age = Variable() \\
  \tt  total\_age = 100 \\
   \tt paul\_age = john\_age + 10 \\
   \tt peter\_age = paul\_age + john\_age \\
   \tt total\_age = peter\_age + paul\_age + john\_age \\
  \tt  result = peter\_age \\
  \tt  \# We know paul\_age = john\_age + 10, and peter\_age = paul\_age + john\_age. So peter\_age = john\_age + 10 + john\_age = 2 * john\_age + 10 \\
   \tt \# We also know total\_age = peter\_age + paul\_age + john\_age = \textcolor{red}{2 * john\_age + 10 + john\_age + 10} = 3 * john\_age + 20 \\
  \tt  \# Therefore, john\_age = (total\_age - 20) / 3 = (100 - 20) / 3 = 80 / 3 = 26.67 \\
  \tt  \# peter\_age = 2 * john\_age + 10 = 2 * 26.67 + 10 = 53.33 + 10 = 63.33 \\
   \tt \# result = peter\_age = 63.33 \\
   \tt \# The answer is 63.33 \\
\bottomrule
    \end{tabularx}
    \caption{Examples of planning errors made by {\sc CoTSolver} from \gsm{}. Both examples are generated by \ttsmall{code-davinci-002}. In the first example, LLM fails to flip the sign of \texttt{colby\_took}. In the second example, the model drops an occurrence of \texttt{john\_age} when expanding out the formula. In both case, LLM fails to arrange the symbols to derive a chain of numeric equations that gives the correct answer. }
    \label{fig:gsm_cotsolver_err_plan}
\end{figure}

\begin{figure}[h]
    \footnotesize
    \centering
    \begin{tabularx}{\linewidth}{X}
    \toprule
            \multicolumn{1}{c}{\bf \textcolor{red}{Execution Errors} of {\sc CoTSolver}: \clutrr{}} \\
         \midrule
    \# [James] was looking forward to dinner with his granddaughter, [Joanne].\\
    relation(James, Joanne) = (grandfather, granddaughter)\\
    \# [Helen] went to her son [Leo]'s House\\
    relation(Helen, Leo) = (mother, son)\\
    \# [Gene] bought gifts for his lovely sister, [Joanne]\\
    relation(Gene, Joanne) = (brother, sister)\\
    \# [Kecia] wanted to visit an art museum, so she asked her father, [James] to take her.\\
    relation(Kecia, James) = (daughter, father)\\
    \# [Sharon] and her sister, [Kecia], frequently bicker. However, they always make up quickly.\\
    relation(Sharon, Kecia) = (sister, sister)\\
    \# [Leo] was so happy for his brother, [Gene], getting a promotion.\\
    relation(Leo, Gene) = (brother, brother)\\
    \# How is [Sharon] related to [Helen]?\\
    answer = relation(Sharon, Helen)\\
    \# Solve the problem\\
    \# Helen's son is Leo, and Leo's brother is Gene, so Helen's son is Gene.\\
    \# Helen's son is Gene, and Gene's sister is Joanne, so Helen's daughter is Joanne.\\
    \# Helen's daughter is Joanne, and Joanne's father is James, \textcolor{red}{so Helen's son is James.}\\
    \# Helen's son is James, and James's daughter is Kecia, so Helen's granddaughter is Kecia.\\
    \# Helen's granddaughter is Kecia, and Kecia's sister is Sharon, so Helen's granddaughter is Sharon.\\
    \# The answer is granddaughter.\\
    \midrule
       \multicolumn{1}{c}{\bf \textcolor{red}{Planning Errors} of {\sc CoTSolver}:  \clutrr{}}   \\
         \midrule
    \# [Rosa] took her daughter [Melissa] to the park.\\
    relation(Rosa, Melissa) = (mother, daughter)\\
    \# [Melissa] was sad her sister [Annie] could not join.\\
    relation(Melissa, Annie) = (sister, sister)\\
    \# [Lisa] is having a great day at the playground with her mother [Rosa] and [Rosa]'s sister [Ruth].\\
    relation(Lisa, Rosa) = (daughter, mother)\\
    relation(Rosa, Ruth) = (sister, sister)\\
    \# [Rosa] asked her daughter, [Lisa], if she would like to go to a movie with her on Saturday night.\\
    relation(Rosa, Lisa) = (mother, daughter)\\
    \# How is [Ruth] related to [Annie]?\\
    answer = relation(Ruth, Annie)\\
    \# Solve the problem\\
    \# \textcolor{red}{Rosa's daughter is Lisa, and Lisa's mother is Rosa, so Rosa's mother is Rosa.}\\
    \# Rosa's mother is Rosa, and Rosa's sister is Ruth, so Rosa's aunt is Ruth.\\
    \# Rosa's aunt is Ruth, and Rosa's daughter is Melissa, so Rosa's niece is Melissa.\\
    \# Rosa's niece is Melissa, and Melissa's sister is Annie, so Rosa's niece is Annie.\\
    \# The answer is niece.\\
\bottomrule
    \end{tabularx}
    \caption{Examples of planning errors made by {\sc CoTSolver} on \clutrr{}. We omit questions for brevity. Both examples are generated by \ttsmall{code-davinci-002}. In the first example, the model outputs an incorrect value when applying the transitivity rule marked in \textcolor{red}{red} (correct output should be \texttt{husband}). In the second example, the model comes up with an incorrect procedure.  }
    \label{fig:clutrr_cotsolver_err}
\end{figure}

\newpage
\textcolor{white}{something}

\newpage
\textcolor{white}{something}
\newpage

\section{Prompt Examples}
\label{app:detailed_prompts}

We show one or two exemplars in the prompt for each dataset. We list prompts for \pallm{} for comparison.

\textbf{Prompts for \gsm{} and \gsmsys{}}

\begin{figure}[h]
    \footnotesize
    \centering
    \begin{tabularx}{\linewidth}{X}
    \toprule
            \multicolumn{1}{c}{\textbf{\satlm{}} } \\
         \midrule

Q: Jason had 20 lollipops. He gave Denny some lollipops. Now Jason has 12 lollipops. How many lollipops did Jason give to Denny?

\\
 \tt   jason\_lollipops\_initial = 20 \\
  \tt  lollipops\_given = Variable() \\
  \tt  jason\_lollipops\_after = 12 \\
  \tt  jason\_lollipops\_after = jason\_lollipops\_initial - lollipops\_given \\
 \tt   result = lollipops\_given\\
  \tt  solve(result) \\
\\
\\
Q: Jeff bought 6 pairs of shoes and 4 jerseys for \$560. Jerseys cost 1/4 price of one pair of shoes. Find the shoe's price total price.
\\
\\
  \tt  shoes\_num = 6 \\
 \tt   jerseys\_num = 4 \\
  \tt  total\_cost = 560 \\
  \tt  shoes\_cost\_each = Variable() \\
  \tt  jerseys\_cost\_each = Variable() \\
  \tt  shoes\_cost\_each * shoes\_num + jerseys\_cost\_each * jerseys\_num = total\_cost  \\
  \tt  jerseys\_cost\_each = shoes\_cost\_each * 1 / 4 \\
  \tt  shoes\_cost\_total = shoes\_cost\_each * shoes\_num \\
  \tt  result = shoes\_cost\_total \\
  \tt  solve(result) \\
    \midrule
       \multicolumn{1}{c}{{\bf \pallm{}} from \cite{pal} }   \\
   \midrule
Q: Jason had 20 lollipops. He gave Denny some lollipops. Now Jason has 12 lollipops. How many lollipops did Jason give to Denny? \\
\\
\tt    jason\_lollipops\_initial = 20 \\
\tt    jason\_lollipops\_after = 12 \\
 \tt   denny\_lollipops = jason\_lollipops\_initial - jason\_lollipops\_after \\
 \tt   result = denny\_lollipops \\
  \tt  return result \\
\\
\\
Q: Jeff bought 6 pairs of shoes and 4 jerseys for \$560. Jerseys cost 1/4 price of one pair of shoes. Find the shoe's price total price. \\
\\
\tt    shoes\_num = 6 \\
 \tt   jerseys\_num = 4 \\
 \tt   total\_cost = 560 \\
 \tt   jersey\_shoes\_cost\_ratio = 1 / 4 \\
 \tt   shoes\_cost\_each = total\_cost / (shoes\_num + jerseys\_num * jersey\_shoes\_cost\_ratio) \\

\tt    shoes\_cost\_total = shoes\_cost\_each * shoes\_num \\
\tt    result = shoes\_cost\_total \\
\tt    return result \\
\bottomrule
    \end{tabularx}
    \caption{Prompt (excerpt) used for \gsm{} and \gsmsys{}.  }
\end{figure}

\newpage
\textbf{Prompts for \lsat{}}

\begin{figure}[h]
    \footnotesize
    \centering
    \begin{tabularx}{\linewidth}{X}
    \toprule
            \multicolumn{1}{c}{\textbf{\satlm{}} } \\
         \midrule
Nine different treatments are available for a certain illness: three antibiotics—F, G, and H—three dietary regimens—M, N, and O—and three physical therapies—U, V, and W. For each case of the illness, a doctor will prescribe exactly five of the treatments, in accordance with the following conditions: If two of the antibiotics are prescribed, the remaining antibiotic cannot be prescribed. There must be exactly one dietary regimen prescribed. If O is not prescribed, F cannot be prescribed. If W is prescribed, F cannot be prescribed. G cannot be prescribed if both N and U are prescribed. V cannot be prescribed unless both H and M are prescribed. \\
Question: If O is prescribed for a given case, which one of the following is a pair of treatments both of which must also be prescribed for that case?\\
(A) F, M (B) G, V (C) N, U (D) U, V (E) U, W \\
\\

\tt treatments = [F, G, H, M, N, O, U, V, W] \\
\tt antibiotics = [F, G, H] \\
\tt dietary\_regimens = [M, N, O] \\
\tt physical\_therapies = [U, V, W] \\
\tt prescribed = Function(treatments, bool) \\

\tt Count([t:treatments], prescribed(t)) == 5 \\
\tt Count([a:antibiotics], prescribed(a)) <= 2 \\
\tt Count([d:dietary\_regimens], prescribed(d)) == 1 \\
\tt Implies(Not(prescribed(O)), Not(prescribed(F))) \\
\tt Implies(prescribed(W), Not(prescribed(F))) \\
\tt Implies(And(prescribed(N), prescribed(U)), Not(prescribed(G))) \\
\tt Implies(prescribed(V), And(prescribed(H), prescribed(M))) \\
\\
\tt solve(Implies(prescribed(O), And(prescribed(U), prescribed(V)))) \# (A) \\
\tt solve(Implies(prescribed(O), And(prescribed(G), prescribed(V)))) \# (B) \\
\tt solve(Implies(prescribed(O), And(prescribed(N), prescribed(U)))) \# (C) \\
\tt solve(Implies(prescribed(O), And(prescribed(U), prescribed(V)))) \# (D) \\
\tt solve(Implies(prescribed(O), And(prescribed(U), prescribed(W)))) \# (E)\\

    \midrule
       \multicolumn{1}{c}{ {\sc \textbf{CoT}} (annotated by our authors) }   \\
       \midrule
 Nine different treatments are available for a certain illness: three antibiotics—F, G, and H—three dietary regimens—M, N, and O—and three physical therapies—U, V, and W ...... (omitted, see above) \\
Question: If O is prescribed for a given case, which one of the following is a pair of treatments both of which must also be prescribed for that case?\\
(A) F, M (B) G, V (C) N, U (D) U, V (E) U, W \\
\\
\tt Let's first analyze the conditions. We know there must be exactly one dietary regimen prescribed. If O is a prescribed dietary regimen, then M and N must not be prescribed. We know V cannot be prescribed unless both H and M are prescribed, so V cannot be prescribed.\\
\tt Let's consider option (A). M must not be prescribed. This option is False.\\
\tt Let's consider option (B). V cannot be prescribed. This option is False.\\
\tt Let's consider option (C). N must not be prescribed. This option is False.\\
\tt Let's consider option (D). V cannot be prescribed. This option is False.\\
\tt Let's consider option (E). Let's assume U is not prescribed. Because exactly 5 of the treatments will be prescribed, they must be F, G, H, O, and W. In this way both F and W are prescribed, which contradicts the condition that if W is prescribed, F cannot be prescribed. So U must be prescribed. We can prescribe G, H, O, U, W. This option is True.\\
\tt So the answer is (E).\\

\bottomrule
    \end{tabularx}
    \caption{Prompt (excerpt) used for \lsat{}. Another example can be found in Figure~\ref{fig:framework}. Several of our authors authored different CoT prompts, leading to similar performance (ranging from 20\% to 22\%). }
\end{figure}

\newpage
\textbf{Prompts for \boardgame{}}

\begin{figure}[h]
    \footnotesize
    \centering
    \begin{tabularx}{\linewidth}{X}
    \toprule
            \multicolumn{1}{c}{\textbf{\satlm{}} } \\
         \midrule
A few players are playing a boardgame. The current state of the game is as follows. The gecko has 13 friends, and hates Chris Ronaldo. And the rules of the game are as follows. Rule1: If the gecko has more than 8 friends, then the gecko does not proceed to the spot that is right after the spot of the bat. Rule2: Regarding the gecko, if it is a fan of Chris Ronaldo, then we can conclude that it does not proceed to the spot that is right after the spot of the bat. Rule3: If something does not proceed to the spot right after the bat, then it does not give a magnifier to the swordfish. \\
Q: Based on the game state and the rules and preferences, does the gecko give a magnifier to the swordfish?\\
\\
\tt    \# If the gecko has more than 8 friends, then the gecko does not proceed to the spot that is right after the spot of the bat. \\
\tt   Implies(has\_more\_than\_8\_friends(gecko), Not(proceed\_to\_spot\_right\_after(gecko, bat))) \\
\tt    \# Rule2: Regarding the gecko, if it is a fan of Chris Ronaldo, then we can conclude that it does not proceed to the spot that is right after the spot of the bat. \\
\tt Implies(is\_fan\_of\_chris\_ronaldo(gecko), Not(proceed\_to\_spot\_right\_after(gecko, bat)))\\
\tt    \# Rule3: If something does not proceed to the spot right after the bat, then it does not give a magnifier to the swordfish.\\
 \tt   ForAll([x], Implies(Not(proceed\_to\_spot\_right\_after(x, bat)), Not(give\_magnifier(x, swordfish))))\\
\\
 \tt   \# The current state of the game is as follows. The gecko has 13 friends, and hates Chris Ronaldo.\\
\tt    \# The gecko has 13 friends.\\
\tt    has\_more\_than\_8\_friends(gecko) == True\\
\tt    \# The gecko hates Chris Ronaldo.\\
\tt    is\_fan\_of\_chris\_ronaldo(gecko) == False\\
\\
 \tt   \# question: does the gecko give a magnifier to the swordfish? \\
\tt    solve(give\_magnifier(gecko, swordfish)) \\

    \midrule
       \multicolumn{1}{c}{ {\sc \textbf{CoT}} from \citet{boardgameqa} }   \\
       \midrule
A few players are playing a boardgame. The current state of the game is as follows. The gecko has 13 friends, and hates Chris Ronaldo. And the rules of the game are as follows. Rule1: If the gecko has more than 8 friends, then the gecko does not proceed to the spot that is right after the spot of the bat. Rule2: Regarding the gecko, if it is a fan of Chris Ronaldo, then we can conclude that it does not proceed to the spot that is right after the spot of the bat. Rule3: If something does not proceed to the spot right after the bat, then it does not give a magnifier to the swordfish. \\
Q: Based on the game state and the rules and preferences, does the gecko give a magnifier to the swordfish?\\
\\
A: We know the gecko has 13 friends, 13 is more than 8, and according to Rule1 "if the gecko has more than 8 friends, then the gecko does not proceed to the spot right after the bat", so we can conclude "the gecko does not proceed to the spot right after the bat". We know the gecko does not proceed to the spot right after the bat, and according to Rule3 "if something does not proceed to the spot right after the bat, then it doesn't give a magnifier to the swordfish", so we can conclude "the gecko does not give a magnifier to the swordfish". So the statement "the gecko gives a magnifier to the swordfish" is disproved. The answer is no.\\
\bottomrule
\end{tabularx}
\caption{Prompt (excerpt) used for \boardgame{}{}. }
\end{figure}

\newpage

\textbf{Prompts for \clutrr{}}
\begin{figure}[h]
    \footnotesize
    \centering
    \begin{tabularx}{\linewidth}{X}
    \toprule
            \multicolumn{1}{c}{\textbf{\satlm{}} } \\
         \midrule
Dorothy took her daughter Michelle and her mother Gabrielle car shopping.\\
Q: How is [Michelle] related to [Gabrielle]? \\
  \tt  \# [Dorothy] took her daughter [Michelle] and her mother [Gabrielle] car shopping.\\
  \tt  relation(Dorothy, Michelle) = (mother, daughter)\\
  \tt  relation(Dorothy, Gabrielle) = (daughter, mother)\\
   \tt \# How is [Michelle] related to [Gabrielle]? \\
   \tt solve(relation(Michelle, Gabrielle)) \\
    \\
Teresa and her brother Ellis were having a wonderful time at Disneyland. Ellis asked his grandmother, Molly, to read him a bedtime story. Molly read him Hansel \& Gretel, which the boy always loved. Sandra is married to Thomas, the couple welcomed Teresa into the world. \\
Q: How is [Molly] related to [Sandra]? \\
  \tt  \# [Teresa] and her brother [Ellis] were having a wonderful time at Disneyland. \\
   \tt relation(Teresa, Ellis) = (sister, brother) \\
  \tt  \# [Ellis] asked his grandmother, [Molly], to read him a bedtime story. \\
  \tt  relation(Ellis, Molly) = (grandson, grandmother) \\
  \tt  \# [Sandra] is married to Thomas, the couple welcomed [Teresa] into the world.\\
   \tt relation(Sandra, Teresa) = (mother, daughter) \\
  \tt  \# How is [Molly] related to [Sandra]? \\
  \tt  solve (relation(Molly, Sandra))\\
  \midrule
    \multicolumn{1}{c}{\textbf{\pallm{}} from \cite{faithfulcot}} \\
         \midrule
Dorothy took her daughter Michelle and her mother Gabrielle car shopping. \\
Q: How is [Michelle] related to [Gabrielle]? \\
\tt \# To answer this question, we write a program to answer the following subquestions: \\
\tt \# 1. How is [Michelle] related to [Dorothy]? (independent, support: "[Dorothy] took her daughter [Michelle] and her mother [Gabrielle] car shopping.") \\
\tt relation(Michelle, Dorothy) = daughter\\
\tt \# 2. How is [Dorothy] related to [Gabrielle]? (independent, support: "[Dorothy] took her daughter [Michelle] and her mother [Gabrielle] car shopping.") \\
\tt relation(Dorothy, Gabrielle) = daughter \\
\# 3. Final answer: How is [Michelle] related to [Gabrielle]? (depends on 1, 2)\\
\tt relation(Michelle, Gabrielle) = relation(Michelle, Dorothy) @ relation(Dorothy, Gabrielle) \\
\\
Teresa and her brother Ellis were having a wonderful time at Disneyland...... (omitted, see above) \\
Q: How is [Molly] related to [Sandra]? \\
\tt \# To answer this question, we write a program to answer the following subquestions: \
\tt \# 1. How is [Molly] related to [Ellis]? (independent, support: "[Ellis] asked his grandmother, [Molly], to read him a bedtime story.") \\
\tt relation(Molly, Ellis) = grandmother\\
\tt \# 2. How is [Ellis] related to [Teresa]? (independent, support: "[Teresa] and her brother [Ellis] were having a wonderful time at Disneyland.") \\
\tt relation(Ellis, Teresa) = brother\\
\tt \# 3. How is [Teresa] related to [Sandra]? (independent, support: "[Sandra] is married to Thomas, the couple welcomed [Teresa] into the world.")\\
\tt relation(Teresa, Sandra) = daughter \\
\tt \# 4. Final answer: How is [Molly] related to [Sandra]? (depends on 1, 2, 3) \\
\tt relation(Molly, Sandra) = relation(Molly, Ellis) @ relation(Ellis, Teresa) @ relation(Teresa, Sandra) \\
    \bottomrule
    \end{tabularx}
    \caption{Prompt (excerpt) used for \clutrr{}. }
\end{figure}

\newpage
\textbf{Prompts for \proofwriter{}}
\begin{figure}[h]
    \footnotesize
    \centering
    \begin{tabularx}{\linewidth}{X}
    \toprule
            \multicolumn{1}{c}{\textbf{\satlm{}} } \\
         \midrule

Here are some facts and rules:\\
If someone visits the squirrel and the squirrel visits the rabbit then they are round.
All round people are not kind.
If someone is round then they chase the rabbit.
If someone is red and they chase the rabbit then they visit the dog.
If someone is red then they visit the squirrel.
If someone visits the squirrel then the squirrel visits the rabbit.
the rabbit visits the dog. \\
the squirrel chases the bald eagle.
the squirrel chases the rabbit.
the dog sees the bald eagle.  
the bald eagle does not chase the dog.
the bald eagle is red.
the squirrel is round.
the rabbit does not see the dog.
the rabbit sees the bald eagle.
the rabbit sees the squirrel.
the dog does not see the rabbit.
the rabbit does not visit the bald eagle.
the dog does not chase the bald eagle. \\
Q: The statement "The bald eagle visits the dog" is True or False? \\
\\
  \tt  ForAll([x], Implies(And(visit(x, squirrel), visit(squirrel, rabbit)), round(x))) \\
   \tt ForAll([x], Implies(round(x), Not(kind(x)))) \\
   \tt  ForAll([x], Implies(round(x), chase(x, rabbit))) \\
    \tt ForAll([x], Implies(And(red(x), chase(x, rabbit)), visit(x, dog)))\\
    \tt ForAll([x], Implies(red(x), visit(x, squirrel))) \\
    \tt ForAll([x], Implies(visit(x, squirrel), visit(squirrel, rabbit)))  \\
  \tt  chase(squirrel, rabbit) \\
    \tt see(dog, bald\_eagle) \\
    \tt Not(chase(bald\_eagle, dog)) \\
    \tt red(bald\_eagle) \\
    \tt round(squirrel) \\
    \tt Not(see(rabbit, dog)) \\
    \tt see(rabbit, bald\_eagle) \\
    \tt see(rabbit, squirrel) \\
    \tt Not(see(dog, rabbit)) \\
    \tt Not(visit(rabbit, bald\_eagle)) \\
    \tt Not(chase(dog, bald\_eagle))\\
\\
    \tt solve(visit(bald\_eagle, dog)) \\
    \midrule
    \multicolumn{1}{c}{\textbf{\pallm{}} adapted from \cite{creswell2023selectioninference}} \\
    \midrule
             Here are some facts and rules:\\
If someone visits the squirrel and the squirrel visits the rabbit then they are round...... (omitted, see above) \\
Q: The statement "The bald eagle visits the dog" is True or False? \\
\\
    \tt \# the bald eagle is red. \\
    \tt bald\_eagle\_is\_red = True \\
    \tt \# If someone is red then they visit the squirrel. \\
    \tt bald\_eagle\_visits\_squirrel = bald\_eagle\_is\_red \\
    \tt \# If someone visits the squirrel then the squirrel visits the rabbit. \\
    \tt squirrel\_visits\_rabbit = bald\_eagle\_visits\_squirrel \\
    \tt \# If someone visits the squirrel and the squirrel visits the rabbit then they are round. \\
    \tt bald\_eagle\_is\_round = bald\_eagle\_visits\_squirrel and squirrel\_visits\_rabbit \\
    \tt \# If someone is round then they chase the rabbit. \\
    \tt bald\_eagle\_chases\_rabbit = bald\_eagle\_is\_round \\
    \tt \# If someone is red and they chase the rabbit then they \tt visit the dog. \\
    \tt bald\_eagle\_visits\_dog = bald\_eagle\_is\_red and \tt bald\_eagle\_chases\_rabbit \\
    \tt \# Question: The statement "The bald eagle visits the dog" is True or False? \\
    \tt return bald\_eagle\_visits\_dog \\
    
    \bottomrule
    \end{tabularx}
    \caption{Prompt (excerpt) used for \proofwriter{}. }
\end{figure}

\newpage
\textbf{Prompts for \textsc{ColoredObject}}
\begin{figure}[h]
    \footnotesize
    \centering
    \begin{tabularx}{\linewidth}{X}
    \toprule
            \multicolumn{1}{c}{\textbf{\satlm{}} } \\
            \midrule
            Q: On the table, you see a bunch of objects arranged in a row: a purple paperclip, a pink stress ball, a brown keychain, a green scrunchiephone charger, a mauve fidget spinner, and a burgundy pen. What is the color of the object directly to the right of the stress ball? \\
\\
\\
\tt \# What is the color of the object directly to the right of the stress ball? \\
\\
\tt stress\_ball = next(x:objects, name(x) == 'stress ball') \\
\\
\tt direct\_right = next(x:objects, index(x) - index(stress\_ball) == 1) \\
\\
\tt solve(color(direct\_right)) \\
         \midrule
\multicolumn{1}{c}{\textbf{\pallm{}} from \cite{pal} } \\
\midrule
Q: On the table, you see a bunch of objects arranged in a row: a purple paperclip, a pink stress ball, a brown keychain, a green scrunchiephone charger, a mauve fidget spinner, and a burgundy pen. What is the color of the object directly to the right of the stress ball? \\
\\
                     
\tt \# Find the index of the stress ball \\
\tt stress\_ball\_idx = None \\
\tt for i, object in enumerate(objects): \\
\tt   \quad  if object[0] == 'stress ball': \\
  \tt \quad \quad stress\_ball\_idx = i \\
        \quad \quad break \\
\\
\\
\tt \# Find the directly right object \\ 
\tt direct\_right = objects[i+1] \\
\\
\\
\tt \# Check the directly right object's color \\
\tt direct\_right\_color = direct\_right[1] \\
\tt answer = direct\_right\_color\\
\\
\tt return answer \\
         \bottomrule
    \end{tabularx}
    \caption{Prompt (excerpt) used for {\sc Colored Object}. }
\end{figure}
\newpage
\textbf{Prompts for \textsc{StructuredRegex}}
\begin{figure}[h]
    \footnotesize
    \centering
    \begin{tabularx}{\linewidth}{X}
    \toprule
            \multicolumn{1}{c}{\textbf{\satlm{}} } \\
         \midrule
\\
\\
Find the regex for the described patterns. Each regex r can be composed using sub-regexes r1, r2, r3, ... \\
\\
\\
Pattern:\\
Three strings separated by semicolons. The first string can either be 579 or 719, the second and third are composed by three digits or three lower case letters that can be followed by a lower case letter, a digit or a capital letter.\\
\tt r = concat(r1,concat(<;>,concat(r2,concat(<;>,r2)))) \\
\tt r1 = or(<579>,<719>) \\
\tt r2 = concat(or(r3,r4),optional(r5)) \\
\tt r3 = repeat(<num>,3) \\
\tt r4 = repeat(<low>,3) \\
\tt r5 = or(<low>,or(<num>,<cap>)) \\
\midrule
    \multicolumn{1}{c}{\textbf{\pallm{}} from \cite{pal} } \\
    \midrule
\\
\\
Find the regex for the described patterns.\\
\\
\\
Pattern:\\
Three strings separated by semicolons. The first string can either be 579 or 719, the second and third are composed by three digits or three lower case letters that can be followed by a lower case letter, a digit or a capital letter.\\
Regex:\\
\tt concat(or(<579>,<719>),concat(<;>,concat(concat(or(repeat(<num>,3),repeat(<low>,3)),\\
\tt optional(or(<low>,or(<num>,<cap>)))),concat(<;>,concat(or(repeat(<num>,3),repeat(<low>,3)),\\
\tt optional(or(<low>,or(<num>,<cap>)))))))) \\
         \bottomrule
    \end{tabularx}
    \caption{Prompt (excerpt) used for {\sc StructuredRegex}. }
\end{figure}
\newpage

\textbf{Prompts for \textsc{Sat\textsubscript{CoTSolver}}}
\begin{figure}[h]
    \footnotesize
    \centering
    \begin{tabularx}{\linewidth}{X}
    \toprule
            \multicolumn{1}{c}{\textbf{\textsc{Sat\textsubscript{CoTSolver}} for \gsm{}} } \\
            \midrule
Q: Jason had 20 lollipops. He gave Denny some lollipops. Now Jason has 12 lollipops. How many lollipops did Jason give to Denny?\\
\\

    \tt jason\_lollipops\_initial = 20 \\
    \tt lollipops\_given = Variable() \\
    \tt jason\_lollipops\_after = 12 \\
    \tt jason\_lollipops\_after = jason\_lollipops\_initial - lollipops\_given \\
    \tt result = lollipops\_given \\
    \tt solve(result) \\
    \tt \# Solve the value of result \\
    \tt \# We know jason\_lollipops\_after = jason\_lollipops\_initial - lollipops\_given \\
    \tt \# So lollipops\_given = jason\_lollipops\_initial - jason\_lollipops\_after = 20 - 12 = 8 \\
    \tt \# result = lollipops\_given = 8 \\
    \tt \# The answer is 8 \\
         \midrule
\multicolumn{1}{c}{\textbf{\textsc{SAT\textsubscript{CoTSolver}} for \clutrr{}}} \\
\midrule

Dorothy took her daughter Michelle and her mother Gabrielle car shopping.\\
Q: How is [Michelle] related to [Gabrielle]?\\
   \tt \# [Dorothy] took her daughter [Michelle] and her mother [Gabrielle] car shopping.\\
   \tt relation(Dorothy, Michelle) = (mother, daughter)\\
   \tt relation(Dorothy, Gabrielle) = (daughter, mother)\\
    \tt \# How is [Michelle] related to [Gabrielle]?\\
   \tt solve(relation(Michelle, Gabrielle))\\
   \tt \# Solve the problem \\
    \tt \# Gabrielle's daughter is Dorothy, and Dorothy's daughter is Michelle, so Gabrielle's granddaughter is Michelle.\\
    \tt \# The answer is granddaughter.\\
         \bottomrule
    \end{tabularx}
    \caption{Prompt (excerpt) used for {\sc Sat\textsubscript{CotSolver}}. }
\end{figure}
\end{document}